%% file: aaai25.tex
\documentclass[letterpaper]{article} % DO NOT CHANGE THIS
\usepackage{aaai25}  % DO NOT CHANGE THIS
\usepackage{times}  % DO NOT CHANGE THIS
\usepackage{helvet}  % DO NOT CHANGE THIS
\usepackage{courier}  % DO NOT CHANGE THIS
\usepackage[hyphens]{url}  % DO NOT CHANGE THIS
\usepackage{graphicx} % DO NOT CHANGE THIS
\usepackage{natbib}  % DO NOT CHANGE THIS AND DO NOT ADD ANY OPTIONS TO IT
\usepackage{caption} % DO NOT CHANGE THIS AND DO NOT ADD ANY OPTIONS TO IT
\usepackage{algorithm}
\usepackage{algorithmic}

\usepackage{newfloat}
\usepackage{listings}
\DeclareCaptionStyle{ruled}{labelfont=normalfont,labelsep=colon,strut=off} % DO NOT CHANGE THIS
\floatstyle{ruled}
\newfloat{listing}{tb}{lst}{}
\floatname{listing}{Listing}
\usepackage{svg}
\usepackage{rotating}
\usepackage{makecell}
\usepackage{comment}
\usepackage{caption}
\usepackage{subcaption}
\usepackage{microtype}
\usepackage{inconsolata}
\usepackage[utf8]{inputenc}

\usepackage{latexsym}
\usepackage{amssymb}
\usepackage{amsmath}
\usepackage{amsthm}
\usepackage{booktabs}
\usepackage{enumitem}
\usepackage{graphicx}
\usepackage{color}

\title{The Impact of Model Scaling on Seen and Unseen Language Performance}
\author{
    Rhitabrat Pokharel, Sina Bagheri Nezhad, Ameeta Agrawal, Suresh Singh\\
}
\affiliations{
    Portland State University\\
    Oregon, USA\\
    \{pokharel, sina5, ameeta, singhsp\}@pdx.edu
}

\usepackage{bibentry}
\begin{document}

\maketitle

\begin{abstract}
The rapid advancement of Large Language Models (LLMs), particularly those trained on multilingual corpora, has intensified the need for a deeper understanding of their performance across a diverse range of languages and model sizes. Our research addresses this critical need by studying the performance and scaling behavior of multilingual LLMs in text classification and machine translation tasks across 204 languages. We systematically examine both seen and unseen languages across three model families of varying sizes in zero-shot and few-shot settings. Our findings show significant differences in scaling behavior between zero-shot and two-shot scenarios, with striking disparities in performance between seen and unseen languages. Model scale has little effect on zero-shot performance, which remains mostly flat. However, in two-shot settings, larger models show clear linear improvements in multilingual text classification. For translation tasks, however, only the instruction-tuned model showed clear benefits from scaling. Our analysis also suggests that overall resource levels, not just the proportions of pretraining languages, are better predictors of model performance, shedding light on what drives multilingual LLM effectiveness.
\end{abstract}

\input{AnonymousSubmission/files/Sections/1_intro.tex}

\input{AnonymousSubmission/files/Sections/2_relatedwork.tex}

\input{AnonymousSubmission/files/Sections/3_method.tex}

\input{AnonymousSubmission/files/Sections/4_results.tex}

\input{AnonymousSubmission/files/Sections/6_conclusion.tex}

% \section{Acknowledgments}

\bigskip

\bibliography{aaai25}

\appendix
\section{Appendix}

\input{AnonymousSubmission/files/Sections/7_appendix.tex}

\end{document}

%% file: AnonymousSubmission/files/Sections/1_intro.tex
\section{Introduction}

%Current multilingual Large Language Models (LLMs) are moving towards using as much data from as many languages as possible to unlock exciting new capabilities. Unfortunately, a clear understanding of the behavior of such multilingual models at scale is missing. Scaling laws quantify the relationship between a model’s performance and key design factors such as the size of the training data or the model’s architecture. In the context of LLMs, these laws offer valuable guidance for model development, resource allocation, and selection of appropriate training data. While extensive research has focused on scaling laws in predominantly English tasks and datasets \citep{mckenzie2023inverse, workshop2023bloom, weietal2023inverse, srivastava2023imitation}, their applicability to multilingual contexts necessitates further exploration. 

Current trends in the development of large language models (LLMs)  emphasize the use of extensive data in several different languages and larger model sizes \citep{srivastava2023imitation}. Scaling trends serve to quantify the connection between a model's performance and critical design elements such as training data size, number of model parameters, or architectural intricacies, providing invaluable insights for model refinement, resource distribution, and the selection of pertinent training data. Although significant research efforts have focused on scaling trends within predominantly English settings \citep{radford2019language, lin2022fewshot, mckenzie2023inverse, weietal2023inverse, workshop2023bloom, chiaetal2024instructeval}, their understanding in multilingual contexts remains underexplored \cite{sunmicelibarone2024scaling}.

Several massively multilingual LLMs have been introduced \citep{lin2022fewshot, openai2023gpt4, workshop2023bloom, shliazhko2024mgpt}, however, evaluation is often limited to a few tens of languages, limiting the large-scale evaluation of current multilingual language models in many languages, especially those truly low-resource languages \cite{ahuja-etal-2023-mega}. Moreover, prior work focuses on studying scaling on languages that were {\em seen} during pretraining whereas we extend our focus also to languages that were potentially {\em not seen} during pretraining.

In this paper, we take a step towards performing a large-scale study of the scaling behavior of multilingual models from the lens of multilingual text classification as well as text generation. We look into the relation between model sizes (number of parameters) and the downstream task performance across diverse languages and resource levels. We comprehensively evaluate several different model sizes from three model families across 204 low and high resource languages in two types of tasks -- topic classification and machine translation. Multilingual topic classification was chosen for two key reasons: it is a widely studied and popular task in natural language processing (NLP) \citep{workshop2023bloom}, and the new parallel dataset SIB-200 \cite{adelani2024sib200} enables consistent and unified analysis across 204 languages. Similarly, machine translation with the FLORES-200 dataset allows for comparable evaluation under text generation setting.

Specifically, we assess the performance of \texttt{XGLM} (with model sizes ranging from 564 million to 7.5 billion parameters), \texttt{bloom}, and \texttt{bloomz} models (ranging from 560 million to 7.1 billion parameters). In total, the results of our study are derived from evaluation involving more than 2 million instances. To our knowledge, ours is the first study to investigate the impact of multilingual scaling across more than 200 languages, both seen as well as unseen\footnote{A language is considered as potentially unseen if it is not listed in the language distribution of the model's pretraining data.}, across two different tasks.

We seek to answer the following three questions: {\em 1)} What is the impact of model scaling on seen and unseen languages of different resource levels for each type of task? {\em 2)} How do different settings (zero-shot and few-shot in-context learning) affect the overall model effectiveness? {{\em 3)} What is the correlation between general resource level, the language-specific training data and performance?}

% We seek to answer the following three questions: 
% \begin{enumerate}
%     \item What is the impact of model scaling on seen and unseen languages of different resource levels for each type of task? 
%     \item How do different settings (zero-shot and few-shot in-context learning) affect the overall model effectiveness?
%     \item What is the correlation between general resource level, the language-specific training data and performance?
% \end{enumerate}

\begin{table*}[t]
    \centering
    \setlength{\tabcolsep}{14pt}
    % \small
    \begin{tabular}{p{0.8cm} p{3.9cm} p{1.5cm} p{3.0cm} p{1.0cm}}
    % \begin{tabular}{cccccc}
        \toprule
        \textbf{Model} &\textbf{Model Sizes (\# parameters)} &\textbf{\# Pretrain Tokens} &\textbf{\# Lang. / \# Lang. Families} &\textbf{Vocab.}\\
        \midrule
        \texttt{xglm} & \raggedright 564M, 1.7B, 2.9B, 7.5B &  500B  &30 / 15 &  256,008 \\
        \texttt{bloom} & \raggedright 560M, 1.1B, 1.7B, 3B, 7.1B &  341B &46 / 8 & 250,880 \\
        \texttt{bloomz} & \raggedright 560M, 1.1B, 1.7B, 3B, 7.1B &  341B  &46 / 8 &  250,880\\
        \bottomrule
    \end{tabular}
    \caption{Overview of model details.}
    \label{tab: model_details}
    %\vspace{-0.4cm}
\end{table*}

% \begin{itemize}
%     \item Introduce the significance of LLMs in natural language processing.
%     \item Significance of ICL.
%     \item State the motivation behind studying the scalability of LLMs, emphasizing the importance of model size in handling diverse languages and resource levels.
% \end{itemize}

%{Although large multilingual models can capitalize on the advantages of cross-linguistic transfer learning, the extensive inclusion of multiple languages inevitably diminishes the model's effective capacity for each individual task. This phenomenon, commonly referred to as the "curse of multilinguality", highlights the inherent competition among languages for the finite capacity within the model \cite{conneauetal2020unsupervised}.}

%= rephrase \sina{rephrased in the next paragraph.} -- In this paper, we investigate the performance of xglm (), BLOOM (560m, 1b1, 1b7, 3b, and 7b1), and bloom-z () on a popular NLP task - text classification across 200 languages at the same time. This allows us to make pertinent observations about the impact of parameter size, learning with zero or few instances in a consistent multilingual setting.

Our key observations are as follows:
\begin{itemize}
    \item In multilingual settings, scaling laws do {\em not} apply consistently. Scaling depends on the task and the inference setting -- for example, we see scaling in 2-shot classification tasks but not in zero-shot classification or in zero-shot and 2-shot machine translation.
    \item While larger models consistently leverage the benefits of few-shot ICL for classification tasks for both seen as well as unseen languages, smaller models struggle in few-shot ICL settings instead yielding better results in {\em zero-shot} settings, especially for unseen languages and low resource languages that were seen during the pretraining.
    \item For the generative task, ICL appears to hurt two language models (\texttt{xglm} and \texttt{bloomz}) while only slightly benefiting the third (\texttt{bloom}), regardless of model scale.
    \item For both tasks, we find that performance obtains stronger correlations with general resource levels rather than language-specific training data, which subsequently result in more pronounced disparities in model performance across different resource levels.
\end{itemize}

%\mee{add this paragraph into key findings if needed or delete} Our paper shows that scaling trends for the seen languages do not always hold true for the unseen languages. Our second main finding is that the correlation between performance and general resource availability is stronger than between performance and language-specific pretraining data; and it’s strong for seen languages but not for unseen languages. Third, nuanced analysis of resource levels yielded some unexpected results, where contrary to expectations, languages with minimal resources (0 and 1 level) did not consistently perform the worst in unseen language scenarios. 

%% file: AnonymousSubmission/files/Sections/2_relatedwork.tex
\section{Related Work}
The impact of model scale on language model performance, particularly within English settings, has been extensively explored in recent years with several studies reinforcing the finding that larger models generally result in enhanced performance \cite{radford2019language,hestness2017deep,kaplan2020scaling,rae2022scaling,wei2022emergent,hoffmann2022training,smith2022using,chowdhery2022palm}. Despite the relationship between model scale and performance not always being linear or predictable \cite{wei2022emergent,weietal2023inverse,xia-etal-2023-training}, larger models continue to be developed, including several multilingual large language models such as \texttt{GPT-3} \cite{brown2020language}, \texttt{XGLM} \cite{lin2022fewshot}, \texttt{BLOOM} \cite{workshop2023bloom},  \texttt{LLaMA} \cite{touvron2023llama}, \texttt{PaLM 2} \cite{anil2023palm}, and others. Although massive multilingual models benefit from positive transfer across languages, the performance of a model deteriorates as its language coverage expands, a challenging phenomenon also known as the  ``curse of multilinguality'' \cite{conneauetal2020unsupervised,pfeifferetal2022lifting}. 

%Notably, \citet{smith2022using} found that higher numbers of pretraining tokens contribute to improved accuracy, while \citet{chowdhery2022palm} explored the impact of scale on few-shot learning, emphasizing its significant influence on model performance. However, it's worth noting that \citet{xia-etal-2023-training} demonstrated an inverse scaling phenomenon, indicating complexities in model behaviors, and \citet{srivastava2023imitation} found nuanced relationships between model size and performance metrics.

%\texttt{mBERT} \cite{devlin-etal-2019-bert}, \texttt{XLM-R} \cite{conneau-etal-2020-unsupervised}, \texttt{mBART} \cite{liu-etal-2020-multilingual-denoising}, \texttt{mT5} \cite{xue-etal-2021-mt5}, \texttt{GPT-4} \cite{openai2023gpt4},

Despite the scaling of multilingual models, research on the impact of scaling on performance remains limited. \citet{lin2022fewshot} showed that different tasks lead to different scaling behavior. However, the datasets in their study were limited to at most 15 languages on the same task, and the classification dataset included only one low-resource language. Other prior work has studied scaling experiments for only English tasks \cite{workshop2023bloom, chiaetal2024instructeval} or a limited set of languages \cite{muennighoff-etal-2023-crosslingual,asai2023buffet,yong2022bloom+,winata2022cross,srivastava2023imitation,isik2024scaling,dakle2022understanding, sunmicelibarone2024scaling} or under only zero-shot setting \cite{muennighoff-etal-2023-crosslingual,adelani2024sib200,yong2022bloom+}. \citet{pmlr-v202-fernandes23a} focused on high-resource languages and neural machine translation, observing that scaling multilingual models improves loss irrespective of the proportion of language pairs in the training mixture.

Unlike previous research, we choose to deeply investigate the performance of several multilingual models across 200+ languages, seen and unseen during pretraining, and spanning various resource levels, a distinction not made in prior work. We contribute to this emerging body of research by studying both classification and generation tasks, several model sizes from three different families of open-source models, and across both zero-shot and few-shot settings.

%% file: AnonymousSubmission/files/Sections/3_method.tex
\section{Experimental Setup}
In this section, we present the overall experimental setup including the models considered, evaluation tasks and datasets, seen and unseen languages, and prompting under zero-shot and few-shot settings.

%(finetuning) 3.7B, 0.5B, 8.4B, 8.4B, 4.1B

% \begin{figure}[htbp]
%   \centering
%   \includesvg[width=0.5\textwidth]{figs/0_vs_2_shot.svg}
%   \caption{svg image}
% \end{figure}

\subsection{Models}
We study three different multilingual LLMs -- \texttt{xglm}, \texttt{bloom}, and \texttt{bloomz} --  of varying sizes as shown in Table~\ref{tab: model_details} to investigate their scalability in multilingual task scenarios. Our study spans a total of 14 different models and model sizes.

% \texttt{\textbf{xglm}} is a decoder-only transformer model trained on a corpus covering 30 languages \cite{lin2022fewshot}. 
% \texttt{\textbf{bloom}} is also a decoder-only transformer language model and was trained on the ROOTS corpus which includes 46 natural languages  \cite{workshop2023bloom}. 
% \texttt{\textbf{bloomz}} is a variant of \texttt{bloom} that underwent instruction tuning  (also known as multitask prompted finetuning) using the xP3 dataset which closely follows ROOTS's language distribution to enhance its adaptability across tasks and languages \cite{muennighoff-etal-2023-crosslingual}.

\begin{itemize}
    
\item \texttt{\textbf{xglm}} is a decoder-only transformer model trained on a corpus covering 30 languages \cite{lin2022fewshot}. 
\item \texttt{\textbf{bloom}} is also a decoder-only transformer language model and was trained on the ROOTS corpus which includes 46 natural languages  \cite{workshop2023bloom}. 
\item \texttt{\textbf{bloomz}} is a variant of \texttt{bloom} that underwent instruction tuning  (also known as multitask prompted finetuning) using the xP3 dataset which closely follows ROOTS's language distribution to enhance its adaptability across tasks and languages \cite{muennighoff-etal-2023-crosslingual}. %By finetuning on multilingual tasks with English prompts, BLOOMZ demonstrates improved performance on both English and non-English tasks. 

\end{itemize}

%Additionally, training on machine-translated prompts further enhances its capability to handle diverse linguistic contexts. 
%It was shown to outperform GPT-3 of comparable size in few-shot learning tasks across more than 20 languages.
    
    % \noindent \textbf{Model Sizes:} The XGLM models employed in our experiments had parameter sizes of 564 million, 1.7 billion, 2.9 billion, and 7.5 billion.

    % \noindent \textbf{Model Sizes:} We utilized BLOOM models with parameter sizes of 560 million, 1.1 billion, 1.7 billion, 3 billion, and 7.1 billion.

    % \noindent \textbf{Model Sizes:} BLOOMZ models used in this study has parameter sizes of 560 million, 1.1 billion, 1.7 billion, 3 billion, and 7.1 billion.

Table~\ref{tab:samples} presents some sample instances from the two datasets -- SIB-200 and Flores-200 used in this study.

\begin{table*}[htbp]
    \centering
    %\small
    % \vspace{0.05cm}
    \begin{tabular}{p{14cm}|p{3cm}}
        \toprule
        % \multicolumn{2}{c}{\textit{\textbf{SIB-200 Samples}}} \\
        
        \multicolumn{2}{c}{\textit{SIB-200 Samples}} \\
        \midrule
        \textbf{Text} & \textbf{Category} \\
        \midrule
         \emph{[en]} The fission bomb works on the principle that it takes energy to put together a nucleus with many protons and neutrons. & science/technology \\
        \midrule
        \emph{[fe]} Lors des sélections de 1976, il a conseillé Carter en matière de politique étrangère, puis a été conseiller à la sécurité nationale (NSA) de 1977 à 1981, succédant à Henry Kissinger. & politics \\
        \bottomrule
    \end{tabular}
    % \vspace{0.20cm} \\
    \begin{tabular}{p{8.5cm}|p{8.5cm}}
        %\toprule
        \multicolumn{2}{c}{\textit{Flores-200 Samples}} \\
        \midrule
        \textbf{Text (\emph{xx})} & \textbf{Text (\emph{en})} \\
        \midrule
        \emph{[sw]} Wasafiri wanashauriwa kwa dhati kujua kuhusu hatari yoyote ya anga mbaya inayoathiri eneo lao kwani huenda ikaathiri mipango yoyote ya usafiri. & Travellers are strongly advised to be aware of any risk of severe weather affecting their area as they may affect any travel plans. \\
        \midrule
        \emph{[id]} Potro menerima perawatan untuk bahunya, tetapi mampu kembali ke permainan. & Potro received treatment to his shoulder at this point but managed to return to the game. \\
        \bottomrule
    \end{tabular}
    \caption{Some samples from SIB-200 and Flores-200 datasets.}
    \label{tab:samples}
   
\end{table*}

These models were selected for several reasons. 1) They appear in several sizes with parameter counts ranging from about 500M to 7B, where the different model sizes were still trained on the same pretraining corpus, making them quite comparable across different scales; 2) These models share similar architecture (i.e., they are all decoder-only models). \texttt{xglm} and \texttt{bloom} were pretrained with similar training procedures, while \texttt{bloomz} is an instruction-tuned model, allowing us to study diverse types of models as there is evidence that pretraining setup affects downstream task performance even after instruction-tuning \cite{asai2023buffet}; 3) They can be leveraged via prompt-based interaction without the need for fine-tuning, thus simulating a more natural user experience with LLMs; 
{(4) They include a considerable number of languages in their pretraining data compared to other models like Llama 3 whose pretraining language distribution is unknown;} (5) Most importantly, they are all open-source models with their technical documentation readily available facilitating this investigation.

%We did not utilize the 176B model due to its substantial resource requirements, which exceeded the available resources at our disposal. Given our constraints, focusing on models of smaller sizes enabled us to conduct a more feasible and manageable study while still providing valuable insights into the scalability and performance of Large Language Models across diverse linguistic contexts.

%Previous works have shown that the performance gained by using large model variants over the smaller ones is insignificant compared to the increase in computing resources and time required to train the large model \cite{black-etal-2022-gpt}.

%as this has been extensively studied even in the context of generative models like \texttt{xglm} and \texttt{bloom}/\texttt{bloomz} \cite{lin2022fewshot,workshop2023bloom}, as well as due to its frequent utilization in previous studies to illustrate the scaling behavior of models \cite{mckenzie2023inverse}.

\subsection{Evaluation Tasks, Datasets and Metrics}
We study model performance on both text classification and text generation tasks. Table~\ref{tab:samples} presents some sample instances from the two datasets. For classification, we choose the SIB-200 dataset \cite{adelani2024sib200}, a large-scale benchmark dataset for topic classification across 204 languages and dialects, {7} topics (the labels are: science/technology, travel, politics, sports, health, entertainment and geography), and  21 distinct language families including both high- and low-resource languages.\footnote{It is worth mentioning that while the SIB-200 dataset originally consisted of 204 languages, a recent update added one more language, N'Ko, bringing the total count to 205 languages. However, at the time of our study, we used the version of the dataset  which included 204 languages.} Moreover, SIB-200 dataset is derived from the  Flores-200 machine translation dataset \cite{nllb-22} containing parallel text across multiple languages facilitating direct comparisons. {We evaluate the performance of the models in terms of macro-average F1 score.} 

For text generation, we rely on the Flores-200 dataset \cite{nllb-22} which is a multilingual dataset consisting of parallel text data in 204 different languages, covering a diverse range of language families and regions. We evaluate the performance in terms of SacreBLEU \cite{post-2018-call}. We consider translations in the direction of \emph{xx} to \emph{en} where \emph{xx} is a language from 203 languages (excluding \emph{en}). We focus on the translation direction of \emph{xx} to \emph{en} in this work, as considering the same target language across all scenarios (\emph{en}) allowing verification of the outputs relatively easily for error analysis, yet also effectively assessing the multilingual capabilities of models because source text already encompass 203 languages. The test sets of both datasets include 204 instances per language and both datasets prioritize low-resource languages.

%Introducing additional target languages would add the challenges associated with generating accurate prompts in those languages and deciding whether to utilize \emph{en} exclusively in the prompts. 

% Additionally, generating prompts for translation into languages with specific script types presented a significant challenge. Designing prompts that effectively capture the nuances and complexities of diverse writing systems proved to be a complex task which may affect our evaluation, further motivating our decision to concentrate on English as the target language.

% \textcolor{cyan}{RP: aren't we using SIB200?} \textcolor{red}{SB: Yes, it was because the contents of SIB-200 and Flores-200 are same but it is better to say we tested on Flores-200 as it is a MT dataset.}

%\textit{Resource Levels}: SIB-200 aims to address the lack of evaluation datasets for low-resource languages by including a diverse set of languages from regions such as Africa, the Americas, Oceania, and Southeast Asia. Many of these languages have limited linguistic resources and are typically excluded from standard NLU evaluation benchmarks

%\textit{Language Families}: The dataset encompasses languages from various language families, including but not limited to Indo-European, Afro-Asiatic, Sino-Tibetan, Niger-Congo, Austronesian, and Austroasiatic. It also includes languages from under-represented families such as Nilotic and Atlantic-Congo. 

%While \citet{workshop2023bloom} examined the SuperGLUE task, it primarily focuses on English. 

\subsection{Seen and Unseen Languages}
For each model and dataset, we can further categorize the languages as `seen' or `unseen' based on whether they were included during  the pretraining of the models or not. For instance, out of more than 200 languages present in SIB-200 and Flores-200 datasets, \texttt{xglm}, \texttt{bloom}, and \texttt{bloomz} models have seen only 30, 45, and 45  languages, respectively. Considering languages that are included as well as those that are not explicitly included in the pretraining corpus of the models allows for a more comprehensive analysis.

The categorization of seen and unseen languages was performed as follows. The \texttt{bloom} paper \cite{workshop2023bloom} lists the languages in its pretraining ROOTS corpus using ISO 639-3 codes (e.g., English is ‘eng’), whereas the \texttt{xglm} models' Hugging Face page\footnote{\url{https://huggingface.co/facebook/xglm-564M}} specifies the languages in its pretraining corpus based on ISO 639-1 (e.g., English is ‘en’) codes. In the case of both the datasets, SIB-200 and Flores-200, languages are accompanied by their ISO 639-3 codes and script types. The mapping for \texttt{bloom/bloomz} models and the datasets is straightforward as they both document their languages in similar language codes. To map between  \texttt{xglm} and the datasets, we use ISO-639 Python library\footnote{\url{https://pypi.org/project/iso-639/}} to convert ISO 639-3 codes to ISO 639-1. In a handful of cases, more than one dataset language listed with different scripts got mapped to a single ISO 639-1 code. To resolve this, we employed the most common script type of that language to perform the mapping.

\begin{figure*}[t]
    \centering
    \begin{subfigure}[t]{0.5\textwidth}
        \centering
        \includegraphics[width=0.8\textwidth]{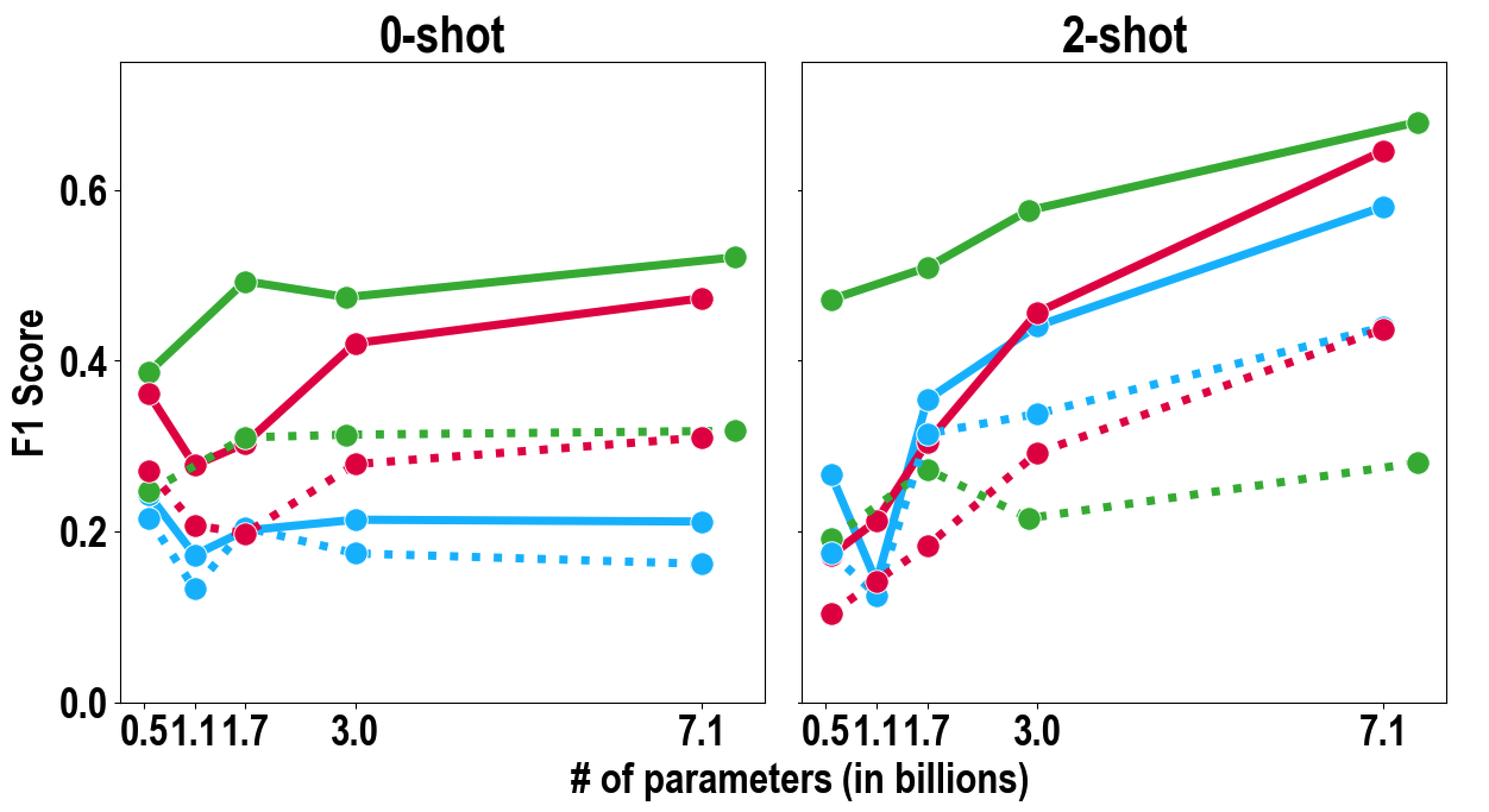}
        \caption{Classification task}
        \label{fig:0_vs_2_class} 
    \end{subfigure}%
    \begin{subfigure}[t]{0.5\textwidth}
        \centering
        \includegraphics[width=1.0\textwidth]{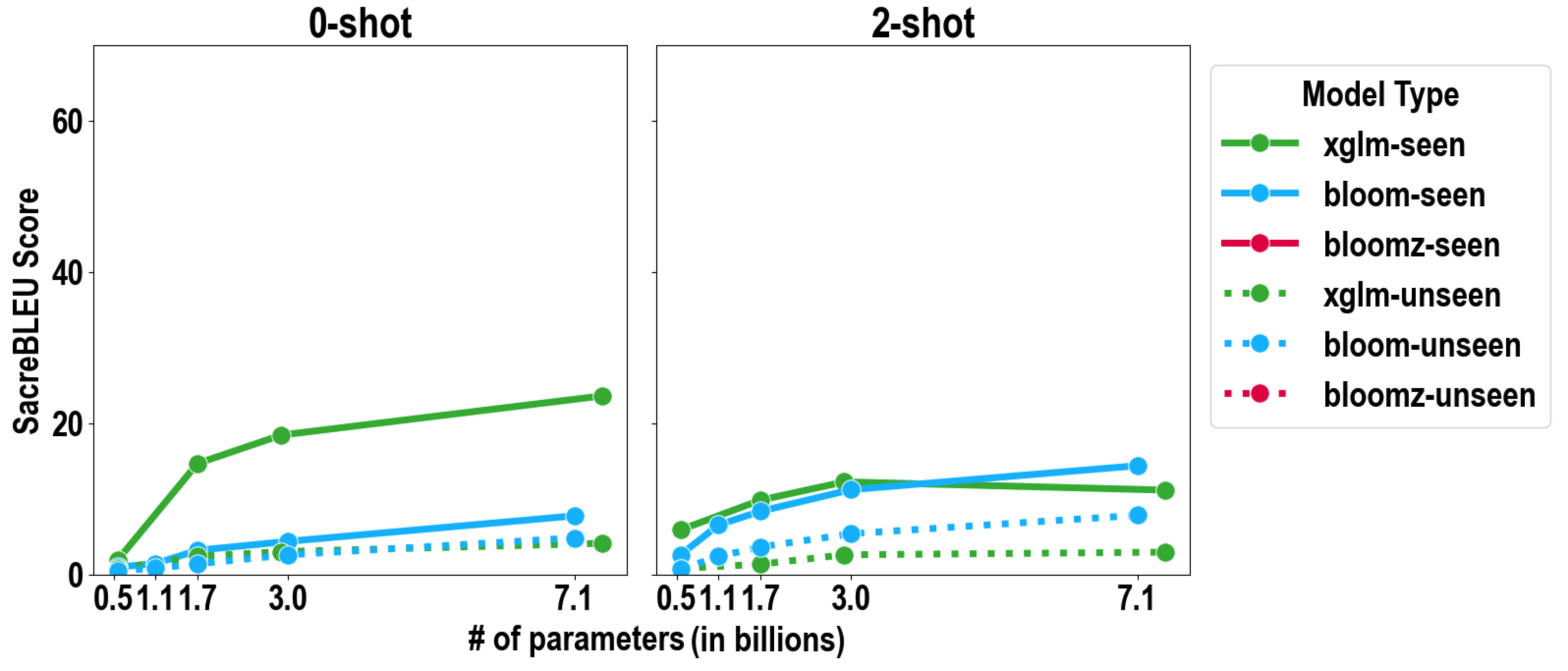}
        \caption{Generation task}
        \label{fig:0_vs_2_gen} 
    \end{subfigure}
    \caption{Multilingual performance under 0-shot and 2-shot settings. Figure \ref{fig:0_vs_2_class} shows results for text classification using SIB-200 topic classification dataset whereas Figure\ref{fig:0_vs_2_gen} shows results for text generation using Flores-200 machine translation dataset. The x-axis shows the model sizes (number of parameters in billions). The y-axis shows F1 scores (0-1) for classification task and SacreBLEU (0-100) for generation task. `\texttt{seen}' indicates the languages that were present in the pretraining data mix of the models whereas `\texttt{unseen}' indicates languages that were not seen by the models during pretraining.}
    \label{fig:zero_vs_few}
\end{figure*}

\subsection{Prompts, Zero-shot, and Few-shot In-context Learning}
%Prompt-based interaction with LLMs has become common not only because of the ease of use and no need for dedicated and expensive resources to fine-tune the models \citep{laietal2023chatgpt}, but also because of the ability of the LLMs to perform tasks in a zero/few-shot setting  \cite{radford2019language}. As such, following prior work \cite{workshop2023bloom}, we focus our evaluation on zero-shot and few-shot ICL settings. % which also reflect the way the models are likely to be used in practice.   

Under \textbf{zero-shot evaluations}, the LLMs were prompted to classify text samples into predefined topic categories (in the case of  SIB-200 dataset) or translate given sentences into English (for Flores-200) without any demonstrations. Under \textbf{few-shot evaluations}, following previous studies \cite{srivastava2023imitation,xia-etal-2023-training} the models were provided with a prompt and a limited number of demonstrations ($k$ = 2).\footnote{Limited context window of some of the models we consider limits the number of $k$ we can consider in our study. For instance, \texttt{xglm} supports 2048 context window length (number of tokens) and as the $n$-shots increase, the text inputs, especially for some languages with high fragmentation during tokenization, become longer than 2048 tokens.} For topic classification, this includes 2 instances per topic class  in the same language as the target language, whereas for machine translation, this includes 2 source-target instances. It is important to note that the same randomly chosen samples from the training sets of the individual datasets were used for all the respective experiments, however, the order in which the demonstrations were presented to the model was randomized for each test instance and this was consistent across all languages. %This means that the order for the first instance of all languages was the same, chosen randomly from the training set. %This randomization aimed to ensure a fair assessment of the models' few-shot learning capabilities while maintaining consistency in the evaluation process across different languages. 

Our prompts were informed by prior work \citep{bachetal2022promptsource, sampathkumar2023} and did not undergo any refinement, generally simulating realistic zero-shot or few-shot scenarios that a user might use while using these models. Informed by findings from previous work which demonstrated the effectiveness of English prompts compared to language-specific prompts 
\cite{lin2022fewshot,laietal2023chatgpt,adelani2024sib200,barreiss2024english,etxaniz2023multilingual}, we used English prompts in our experiments.  %More details are available in Appendix \ref{sec:propmts}. 

For obtaining the label in the text classification task, the models' output probabilities for each topic category were used to determine the predicted label, which was chosen based on the maximum log likelihood among a set of specified candidate label strings \cite{workshop2023bloom}.

In total, our evaluation involved 41,616 test instances from SIB-200 (204 sentences across 204 languages), and  41,412 test instances from Flores-200 (204 sentences across 203 languages), with all the combinations of models, model sizes, zero-shot and few-shot settings yielding results for a total of 2,324,784 instances.

%\noindent \textbf{Training Details} \mee{more hyperparameter details needed, greedy decoding, etc. or say they were default. and move this to appendix. } \sina{As we said we used probabilities we have not any hyperparameters as we did not generate anything. (although it is not compleltely correct as for zero-shot bloomz we used generation with beam search (n beam =4))}

%I understand, but whatever little technical details we can provide might be good - also move this to the appendix.

%not needed -- In terms of hardware resources, we primarily utilized the A6000 48GB GPU for most of the models. For larger models employed in few-shot learning, we leveraged the A100 80GB GPU, which offered enhanced computational capabilities to handle the increased complexity of these models and scenarios. This allocation of hardware resources ensured efficient model training and evaluation across varying scales and complexities.

%% file: AnonymousSubmission/files/Sections/4_results.tex
\section{Multilingual Scaling Results and Analysis}
We now present our results and discuss key findings.

\subsection{Scaling trends in classification vs. generation tasks} 

The overall scaling patterns in multilingual evaluation are presented in Figure~\ref{fig:zero_vs_few}. 

\noindent \textbf{Text classification} \quad For the classification task, in Figure \ref{fig:0_vs_2_class} in the 0-shot setting (left), we observe that both \texttt{bloom} and \texttt{bloomz} models show a predominantly U-shaped scaling trend, while \texttt{xglm} shows a relatively flat trajectory with a subtle hint of the ``double-descent'' phenomenon, where performance initially improves, then declines, and then improves again with increased scale \cite{nakkiran2019deep}. These trends remain consistent for seen as well as unseen languages. However, a more zoomed-out view reveals that, surprisingly, the model sizes appear to have minimal impact as the line plots remain largely stable despite scaling from 560M to 7B, an almost 12.5 fold increase. Our results of scaling under 0-shot setting are different from those observed in an earlier study \cite{lin2022fewshot}. While their results derived from a small set of seen languages show normal scaling under 0-shot  condition for \texttt{xglm}, we see only marginal scaling for \texttt{xglm} and \texttt{bloomz} and none for \texttt{bloom}, possibly because our results consider twice as many seen languages (30 languages) as their setup (15 languages), and more importantly, 13 of our seen languages are from mid- to low-resource categories. %Consequently, when downstream task performance remains comparable, the smaller model would undoubtedly be preferable.  

When we go from the 0-shot to the 2-shot setting, Figure \ref{fig:0_vs_2_class} (right), we find that \texttt{bloomz}'s U-shaped scaling has turned to linear scaling, \texttt{xglm}'s faint double-descent scaling turned to gradual linear scaling for seen languages, and \texttt{bloom}'s U-shaped which was earlier followed by flat lines is now followed by linear scaling. Additionally, it is clear that 2-shot ICL brings additional improvements for all models except for \texttt{xglm} unseen. Nonetheless, these results suggest that multilingual scaling trends for text classification task may differ depending on the settings adopted (0- vs. 2-shot). {This is  similar to our analysis of baseline English-only evaluation behavior (as shown in Figure~\ref{fig:eng_lineplots}) where scaling trends improve moving from 0- to 2-shot. 
%

%and an earlier study where the tasks that were U-shaped in 0-shot continued to be U-shaped after 1-shot (evaluation was only on English) \cite{weietal2023inverse}

% \begin{figure}[h]
%     \centering
%     \includegraphics[width=0.9\linewidth]{AnonymousSubmission/files/figs/eng_lineplots.png}
%     \caption{Performance of the models on English-only subset of SIB-200. On the x-axis are the model sizes (parameters in billions).}
%     \vspace{0.5cm}
%     \label{fig:eng_lineplots}
% \end{figure}

% == english-only results discussion ==  Curiously, \texttt{bloom}'s performance in 0-shot setting takes a sharp dive going from 3B to 7B model as does \texttt{bloomz}'s for the two-shot case, albeit to a smaller degree. Overall, it seems that the largest model size does not necessarily yield the best performance in this text classification task. For all 0-shot experiments, performance plateaus (or worsens) after 3B model sizes in 4 out of 6 cases. }

%, as the inference cost is proportional to its size.

%It is particularly striking that the performance plateaus after reaching the 3B parameter threshold for all models. 

%As models go from zero-shot to 2-shot, performance slightly and inconsistently increases.

\begin{figure}[t]
    \centering
    
    \begin{subfigure}[b]{0.24\textwidth}
        \includegraphics[width=\textwidth]{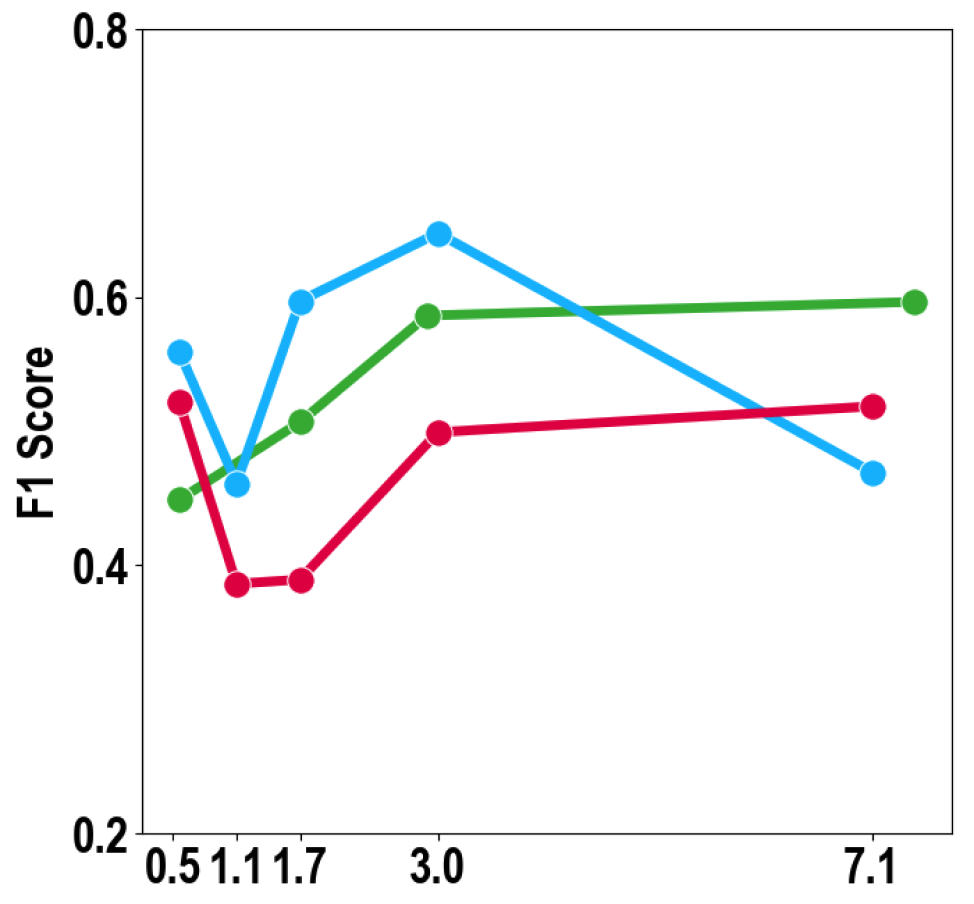}
        \caption{0-shot}
        % \label{fig:figure1}
    \end{subfigure}
    % \hfill
    \begin{subfigure}[b]{0.225\textwidth}
        \includegraphics[width=\textwidth]{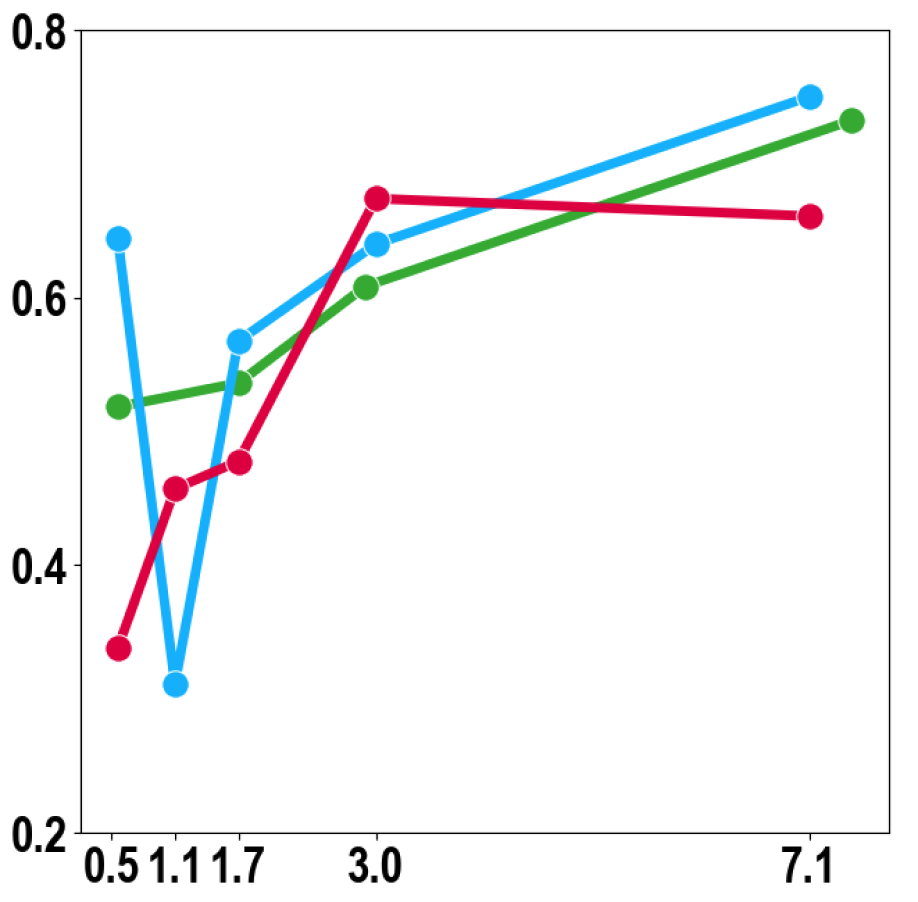}
        \caption{2-shot}
        % \label{fig:figure2}
    \end{subfigure}
    \caption{Performance of three models (\textcolor[HTML]{1DB52C}{\rule{6pt}{6pt}} \texttt{xglm};
            \textcolor[HTML]{0bb4ff}{\rule{6pt}{6pt}} \texttt{bloom};
            \textcolor[HTML]{e60049}{\rule{6pt}{6pt}} \texttt{bloomz}) on English-only subset of SIB-200. On the x-axis are the model sizes. } 
    \label{fig:eng_lineplots}
    % \vspace{-0.3cm}
\end{figure}

\begin{figure}[t]
    \centering
    \includegraphics[width=0.7\linewidth]{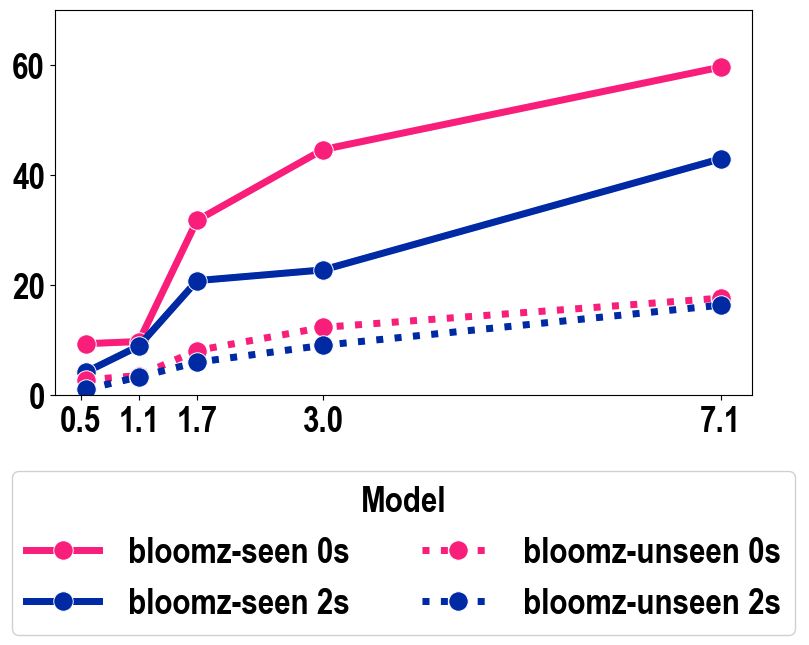}
    \caption{\texttt{bloomz}'s performance on seen and unseen languages for text generation task.}
    % \vspace{-0.2cm}
    \label{fig:bloomz_seen_unseen}
\end{figure}

%\medskip
\noindent \textbf{Text generation} \quad Turning our attention to the generative task, in Figure \ref{fig:zero_vs_few}(b) we observe limited to no scaling for both 0-shot and 2-shot for all cases with the exception of \texttt{xglm} on seen languages in the 0-shot setting\footnote{We do not show the results of \texttt{bloomz} in this plot and instead plot them separately later because part of the Flores-200 dataset was used to instruction-tune bloomz.}. {\em This is in sharp contrast to the text classification task where normal scaling was observed in 2-shot setting}. In fact, we further observe that the performance of \texttt{xglm} degrades when going from 0-shot to the 2-shot case for both seen and unseen languages (similar observations are made for \texttt{bloomz} as shown in Figure~\ref{fig:bloomz_seen_unseen} discussed later). This is also in contrast to the text classification task where few-shot in-context learning brought additional gains for all models. 
{As shown by \cite{zhangetal2023beyond} on QA task and \cite{sunmicelibarone2024scaling} on MT task, scaling behavior varies with prompting methods, task complexity, and model families, which supports the differing results across the two tasks in our experiments.}

\noindent \textbf{Text generation using \texttt{bloomz}} \quad We now analyze the results of \texttt{bloomz} model in text generation under 0-shot and 2-shot settings, as shown in Figure~\ref{fig:bloomz_seen_unseen}. Here we notice normal scaling but a distinctive  {\em degradation} in performance when going from 0-shot to 2-shot setting for both seen and unseen languages. On the one hand, this suggests that instruction-tuned models remain a promising approach to improving the performance of both seen unseen languages with increased scale, although the gap between seen and unseen languages remains significant. On the other hand, our observations also show that while few-shot in-context learning helps in text classification task, it does not seem to be effective in text generation.

\begin{figure*}[t]
    \centering
    \begin{minipage}[t]{0.25\textwidth}
        \centering
        \begin{subfigure}[t]{0.9\textwidth}
            \includegraphics[width=\textwidth]{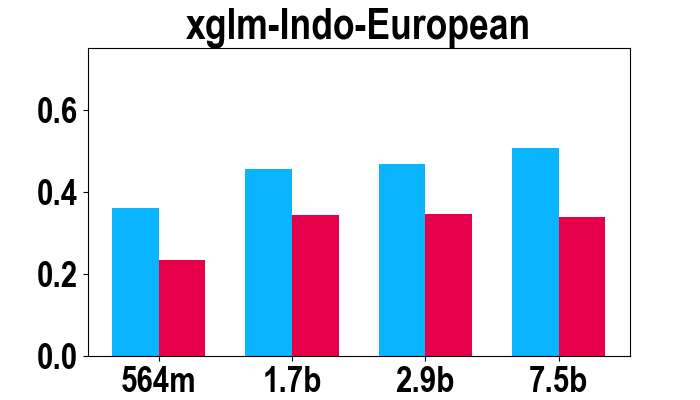}
            % \caption{}
            \label{fig:xglm_Indo-European_class} 
        \end{subfigure}
        \begin{subfigure}[t]{0.9\textwidth}
            \includegraphics[width=\textwidth]{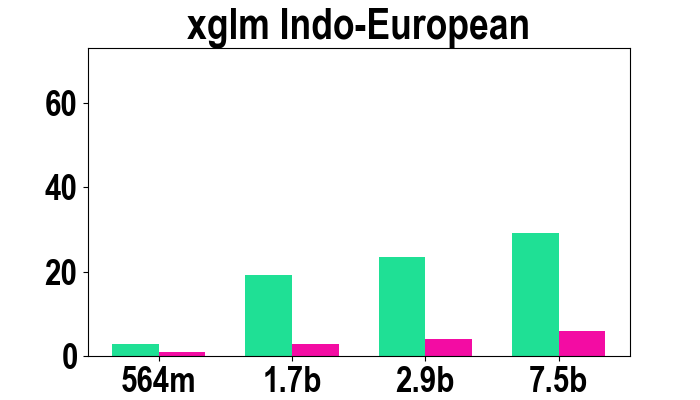}
            % \caption{}
            \label{fig:xglm_Indo-European_gen} 
        \end{subfigure}
    \end{minipage}%
    \begin{minipage}[t]{0.25\textwidth}
        \centering
        \begin{subfigure}[t]{0.9\textwidth}
            \includegraphics[width=\textwidth]{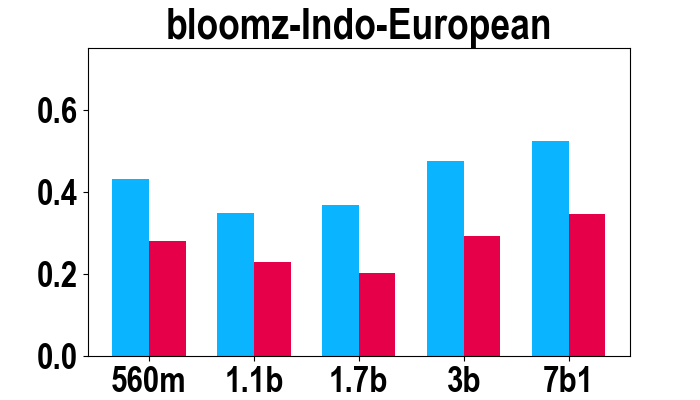}
            % \caption{}
            \label{fig:bloomz_Indo-European_class} 
        \end{subfigure}
        \begin{subfigure}[t]{0.9\textwidth}
            \includegraphics[width=\textwidth]{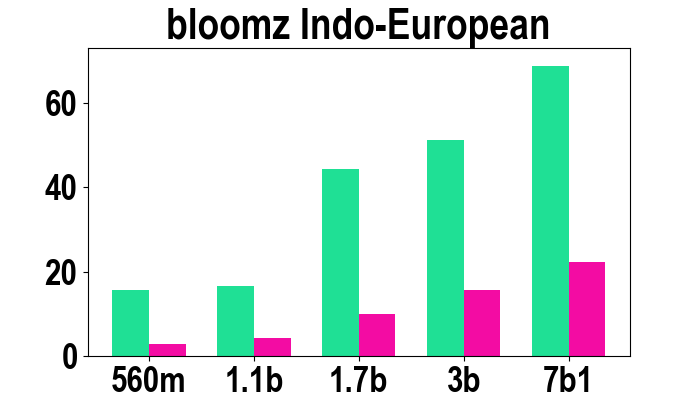}
            % \caption{}
            \label{fig:bloomz_Indo-European_gen} 
        \end{subfigure}
    \end{minipage}%
    \begin{minipage}[t]{0.25\textwidth}
        \centering
        \begin{subfigure}[t]{0.9\textwidth}
            \includegraphics[width=\textwidth]{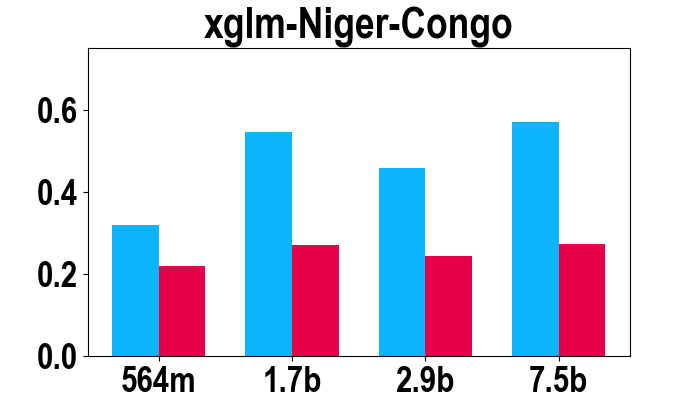}
            % \caption{}
            \label{fig:xglm_Niger-Congo_class}
        \end{subfigure}
        \begin{subfigure}[t]{0.9\textwidth}
            \includegraphics[width=\textwidth]{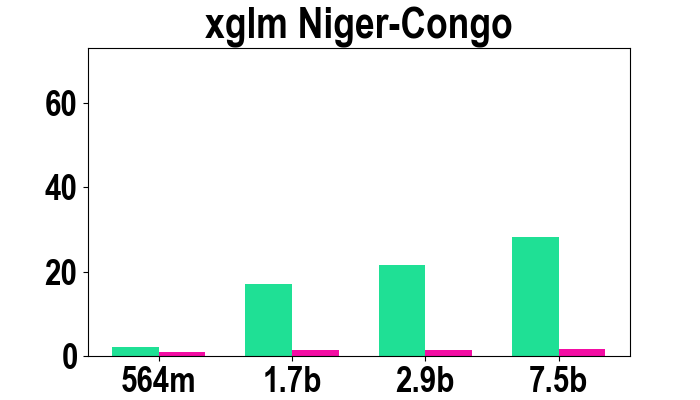}
            % \caption{}
            \label{fig:xglm_Niger-Congo_gen} 
        \end{subfigure}
    \end{minipage}%
    \begin{minipage}[t]{0.25\textwidth}
        \centering
        \begin{subfigure}[t]{0.9\textwidth}
            \includegraphics[width=\textwidth]{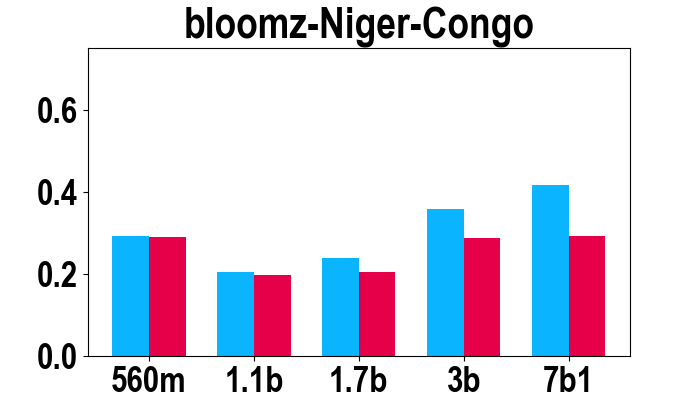}
            % \caption{}
            \label{fig:bloomz_Niger-Congo_class}
        \end{subfigure}
        
        \begin{subfigure}[t]{0.9\textwidth}
            \includegraphics[width=\textwidth]{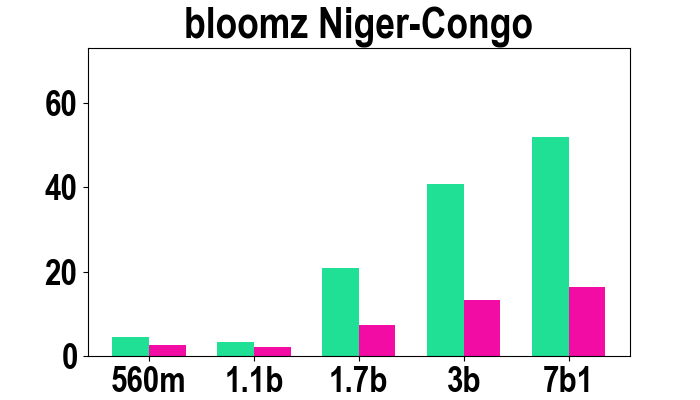}
            % \caption{}
            \label{fig:bloomz_Niger-Congo_gen} 
        \end{subfigure}
    \end{minipage}
    
    \caption{The top row shows performance comparison between seen and unseen for languages on the \textbf{classification task} from the same language family across different model sizes (x-axis) for the 0-shot setting. Similarly, the bottom row shows the comparison for the \textbf{generation task}. \textcolor[HTML]{0bb4ff}{\rule{6pt}{6pt}} \& \textcolor[HTML]{1FE095}{\rule{6pt}{6pt}} - seen; \textcolor[HTML]{e60049}{\rule{6pt}{6pt}} \& \textcolor[HTML]{F30CA3}{\rule{6pt}{6pt}} - unseen.}
    \label{fig:f1_vs_family}
    % \vspace{-0.3cm}
\end{figure*}

%While in the English-only evaluation, there was no clear winner in terms of which model yielded the best results, in the multilingual scenario there is much less debate.

\subsection{Seen vs. unseen languages}  

%We present the results of our analysis focusing on seen and unseen languages in Figure~\ref{fig:f1_vs_family}. 
From the results presented in Figure~\ref{fig:zero_vs_few}, for the \textbf{classification task}, we observe that in both 0-shot and 2-shot scenarios, as expected all models consistently perform better on the `seen' languages compared to  `unseen' languages (solid lines for both tasks), aligning with findings from prior literature \cite{adelani2024sib200}. Specifically, \texttt{xglm} clearly outperforms the other models for `seen' languages, which could be partially explained by a simple statistic -- it encountered fewer languages during pretraining than the \texttt{bloom/bloomz} models, resulting in an average score over a fewer number of languages.

%What is impressive, however, is that \texttt{xglm} also remains competitive for `unseen' languages in the 0-shot setting, despite being trained on a much smaller number of languages than \texttt{bloom}/\texttt{bloomz}. This possibly suggests that simply scaling up the number of seen languages in the training corpus may not necessarily result in improved performance on unseen languages, an observation in line with earlier studies, albeit in slightly different contexts \cite{dakle2022understanding, adelani2024sib200}. 

Two possible explanations for \texttt{xglm}'s comparable performance, despite being trained on a much smaller number of languages than \texttt{bloom}/\texttt{bloomz}, may be found in the fact that 1) it was trained on a much larger number of pretraining tokens (500B) compared to \texttt{bloom}/\texttt{bloomz} models as mentioned earlier in Table~\ref{tab: model_details}, and 2) despite having fewer number of seen languages than \texttt{bloom}/\texttt{bloomz}, \texttt{xglm} has almost twice as many seen {\em language families} as \texttt{bloom}/\texttt{bloomz} which may be potentially contributing to positive cross-lingual transfer. %(more details available in {Supplementary Material}).

On the other hand, in the 2-shot setting, while \texttt{xglm} shows improvement for seen languages, its performance for unseen languages actually worsens. This is in stark contrast to the \texttt{bloom}/\texttt{bloomz} models where 2-shot ICL enhances performance for both seen and unseen languages, and particularly in the case of \texttt{bloom}, quite drastically so. It is not apparent why \texttt{xglm}'s advantage for unseen languages in the 0-shot setting does not carry over in the 2-shot setting. Another intriguing observation is that regardless of whether the languages were seen or unseen during training, \texttt{bloom-1b1} consistently performs at the same level (converging to a specific point) yielding lower performance than its smaller counterpart, \texttt{bloom-560m}.% --  an observation also found by \citet{dakle2022understanding}.  

When we consider the \textbf{generation task}, we see that overall, regardless of 0-shot or 2-shot, the performance of the models is poor. \texttt{bloom} does improve significantly in the 2-shot setting for seen languages as compared to \texttt{xglm} (seen and unseen languages) and \texttt{bloom}-unseen. This behavior is in stark contrast with the classification task where all models and cases (seen/unseen) do better in the 2-shot case. Thus, ICL is useful for classification tasks but not so much for generative tasks.

%We do not include results for \texttt{bloom} because \texttt{bloomz} is fine-tuned and thus displays better performance.

In Figure \ref{fig:f1_vs_family} we dig deeper into the performance of \texttt{xglm} and \texttt{bloomz} for seen and unseen languages divided by language family. {We choose the most highly resourced language family (Indo-European) and one of the least resourced (Niger-Congo).}  In the top row, which corresponds to the \textbf{classification task}, we see that both models show a significant drop when going from seen to unseen languages for a high resource family (Indo-European) as well as for a low resource family (Niger-Congo). In general, the largest model size shows the highest F1 score for both language families. Interestingly, \texttt{bloomz} shows a dip at 1.1b and 1.7b model sizes regardless of language family. 

Moving to the \textbf{generation task} (bottom row in Figure~\ref{fig:f1_vs_family}), we observe a substantial difference in performance between seen and unseen languages for both language families. In this task, text from all languages is translated into \emph{en}. Therefore, the models need to have a good understanding of the input text in order to generate a syntactically and semantically correct translation. We observe that unseen languages are hard to translate, even though the models were trained on other languages from the same language families. However, it is also clear that larger models have developed some ability to understand unseen languages more so than smaller models. For seen languages, we see a clear improvement with increasing model sizes which is due to larger models having greater capacity to retain understanding of these languages. {In contrast, unseen languages benefit significantly less from scaling up the model size.}

\begin{figure*}[th]
    \centering
    \begin{minipage}[t]{0.255\textwidth}
        \centering
        \begin{subfigure}[t]{0.9\textwidth}
            \includegraphics[width=\textwidth]{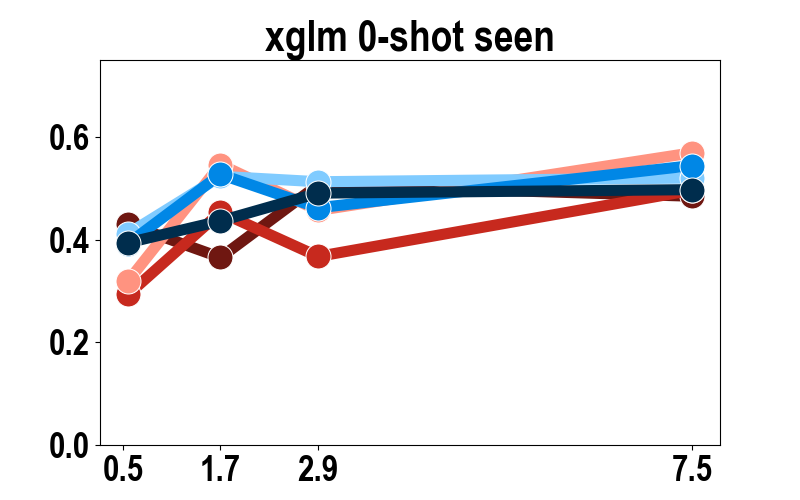}
            % \caption{}
            \label{fig:xglm 0-shot_seen_class}
        \end{subfigure}
    \end{minipage}%
    \begin{minipage}[t]{0.255\textwidth}
        \centering
        \begin{subfigure}[t]{0.9\textwidth}
            \includegraphics[width=\textwidth]{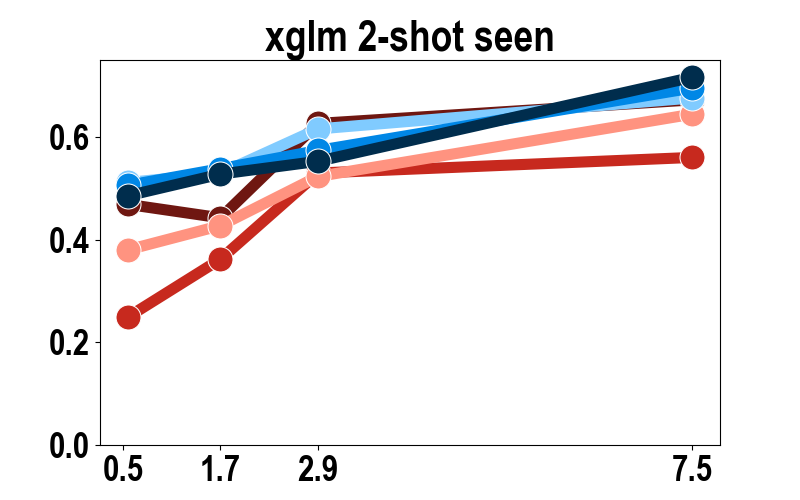}
            % \caption{}
            \label{fig:xglm 2-shot_seen_class}
        \end{subfigure}
    \end{minipage}%
    \begin{minipage}[t]{0.255\textwidth}
        \centering
        \begin{subfigure}[t]{0.9\textwidth}
            \includegraphics[width=\textwidth]{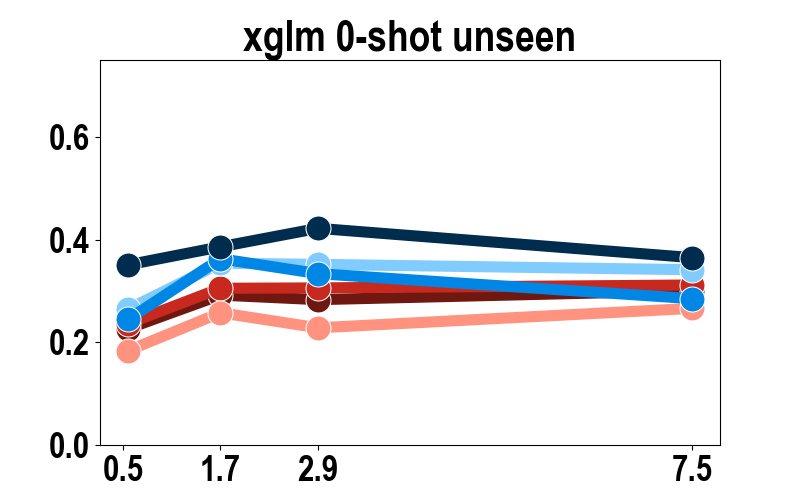}
            % \caption{}
            \label{fig:xglm 0-shot_unseen_class} 
        \end{subfigure}
    \end{minipage}%
    \begin{minipage}[t]{0.255\textwidth}
        \centering
        \begin{subfigure}[t]{0.9\textwidth}
            \includegraphics[width=\textwidth]{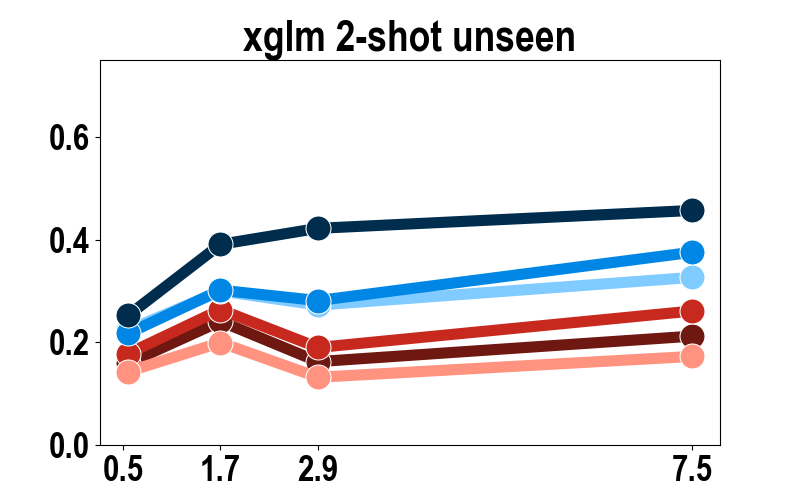}
            % \caption{}
            \label{fig:xglm 2-shot_unseen_class} 
        \end{subfigure}
    \end{minipage}

    \begin{minipage}[t]{0.255\textwidth}
        \centering
        \begin{subfigure}[t]{0.9\textwidth}
            \includegraphics[width=\textwidth]{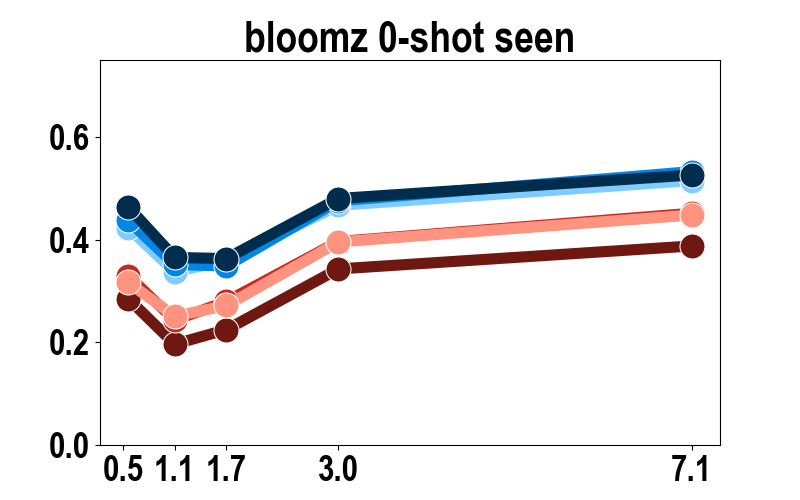}
            % \caption{}
            \label{fig:bloomz 0-shot_seen_class} 
        \end{subfigure}
    \end{minipage}%
    \begin{minipage}[t]{0.255\textwidth}
        \centering
        \begin{subfigure}[t]{0.9\textwidth}
            \includegraphics[width=\textwidth]{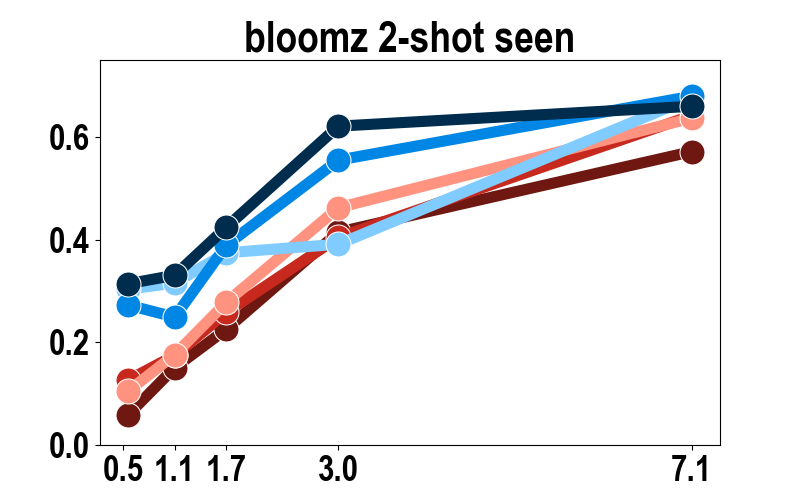}
            % \caption{}
            \label{fig:bloomz 2-shot_seen_class} 
        \end{subfigure}
    \end{minipage}%
    \begin{minipage}[t]{0.255\textwidth}
        \centering
        \begin{subfigure}[t]{0.9\textwidth}
            \includegraphics[width=\textwidth]{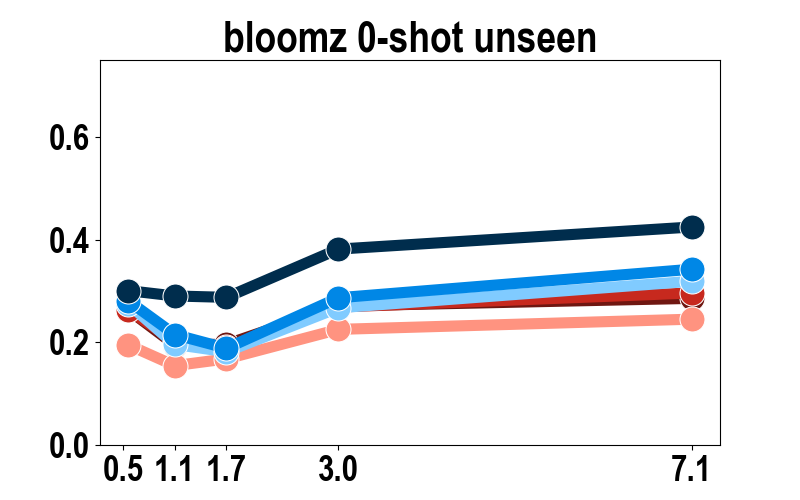}
            % \caption{}
            \label{fig:bloomz 0-shot_unseen_class} 
        \end{subfigure}
    \end{minipage}%
    \begin{minipage}[t]{0.255\textwidth}
        \centering
        \begin{subfigure}[t]{0.9\textwidth}
            \includegraphics[width=\textwidth]{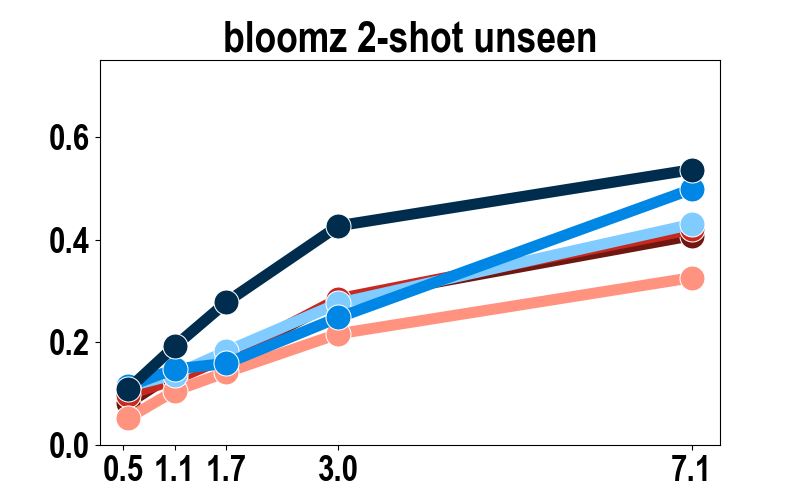}
            % \caption{}
            \label{fig:bloomz 2-shot_unseen_class} 
        \end{subfigure}
    \end{minipage}
    \caption{{\bf Classification task}: Results of evaluation (F1 score) across different models based on language resource level using SIB-200 dataset. The model sizes on the x-axis are in billions of parameters.
            \textcolor[HTML]{6f1711}{\rule{6pt}{6pt}} Resource Level 0, 
            \textcolor[HTML]{c7291e}{\rule{6pt}{6pt}} Resource Level 1, 
            \textcolor[HTML]{ff9380}{\rule{6pt}{6pt}} Resource Level 2, 
            \textcolor[HTML]{80cbff}{\rule{6pt}{6pt}} Resource Level 3, 
            \textcolor[HTML]{0087e6}{\rule{6pt}{6pt}} Resource Level 4, 
            \textcolor[HTML]{002d4d}{\rule{6pt}{6pt}} Resource Level 5.
    }
    \label{fig:lineplots_class}
    %\vspace{-0.5cm}
\end{figure*}

\begin{figure*}[th]
    \centering
    \begin{minipage}[t]{0.255\textwidth}
        \centering
        \begin{subfigure}[t]{0.9\textwidth}
            \includegraphics[width=\textwidth]{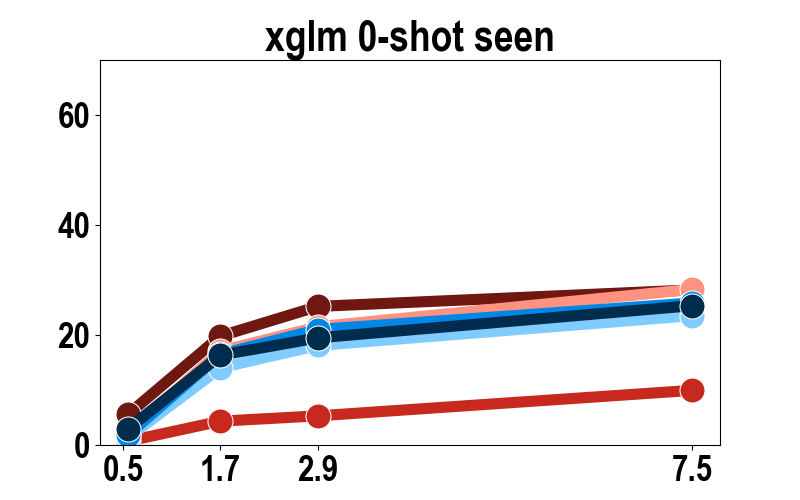}
            % \caption{}
            \label{fig:xglm 0-shot_seen_gen}
        \end{subfigure}
    \end{minipage}%
    \begin{minipage}[t]{0.255\textwidth}
        \centering
        \begin{subfigure}[t]{0.9\textwidth}
            \includegraphics[width=\textwidth]{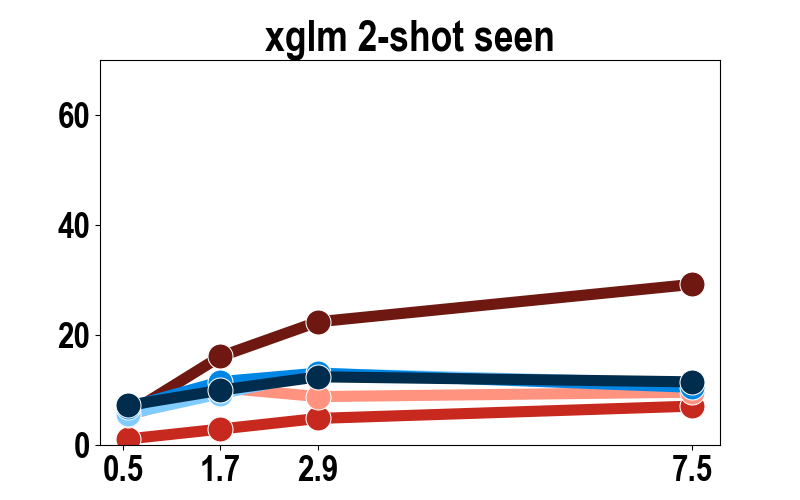}
            % \caption{}
            \label{fig:xglm 2-shot_seen_gen}
        \end{subfigure}
    \end{minipage}%
    \begin{minipage}[t]{0.255\textwidth}
        \centering
        \begin{subfigure}[t]{0.9\textwidth}
            \includegraphics[width=\textwidth]{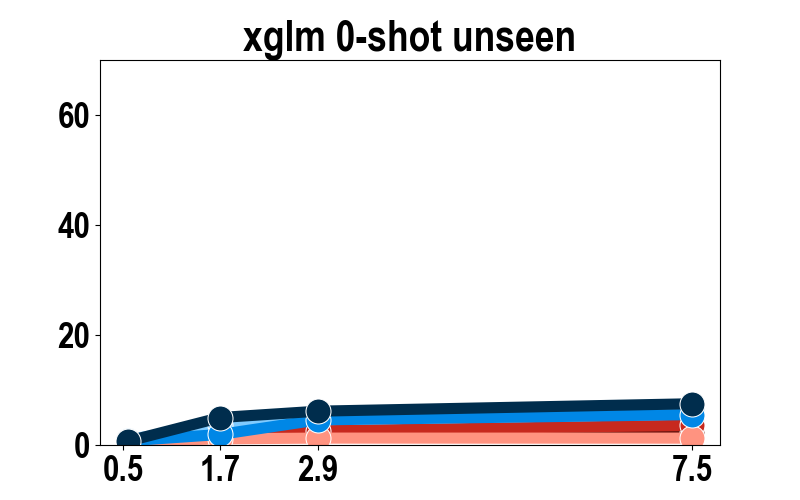}
            % \caption{}
            \label{fig:xglm 0-shot_unseen_gen} 
        \end{subfigure}
    \end{minipage}%
    \begin{minipage}[t]{0.255\textwidth}
        \centering
        \begin{subfigure}[t]{0.9\textwidth}
            \includegraphics[width=\textwidth]{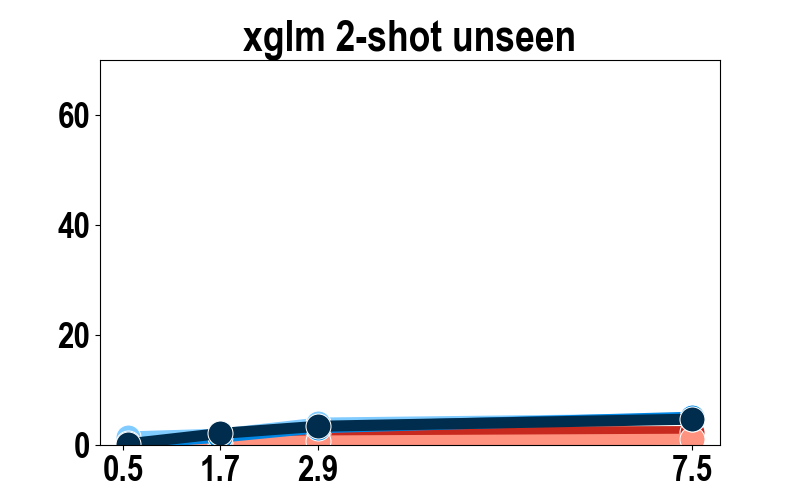}
            % \caption{}
            \label{fig:xglm 2-shot_unseen_gen} 
        \end{subfigure}
    \end{minipage}

     \begin{minipage}[t]{0.255\textwidth}
        \centering
        \begin{subfigure}[t]{0.9\textwidth}
            \includegraphics[width=\textwidth]{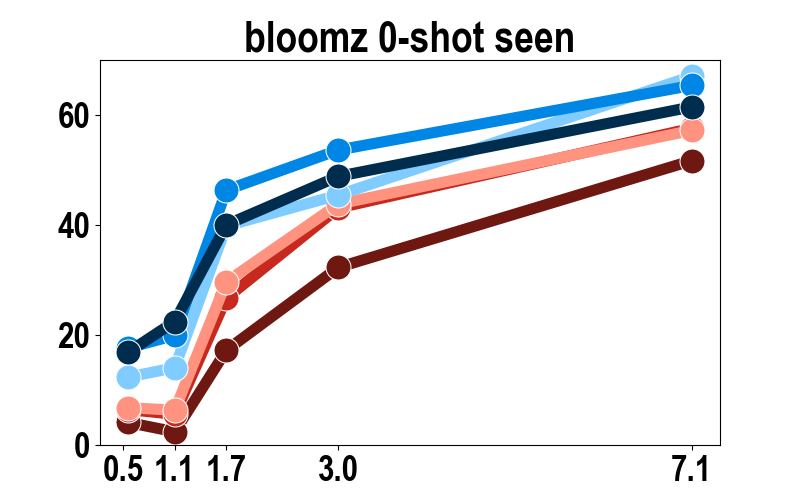}
            % \caption{}
            \label{fig:bloomz 0-shot_seen_gen} 
        \end{subfigure}
    \end{minipage}%
    \begin{minipage}[t]{0.255\textwidth}
        \centering
        \begin{subfigure}[t]{0.9\textwidth}
            \includegraphics[width=\textwidth]{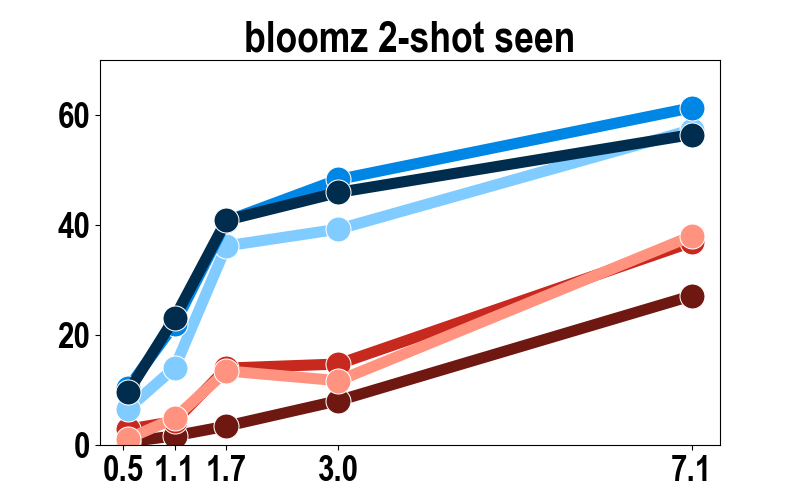}
            % \caption{}
            \label{fig:bloomz 2-shot_seen_gen} 
        \end{subfigure}
    \end{minipage}%
    \begin{minipage}[t]{0.255\textwidth}
        \centering
        \begin{subfigure}[t]{0.9\textwidth}
            \includegraphics[width=\textwidth]{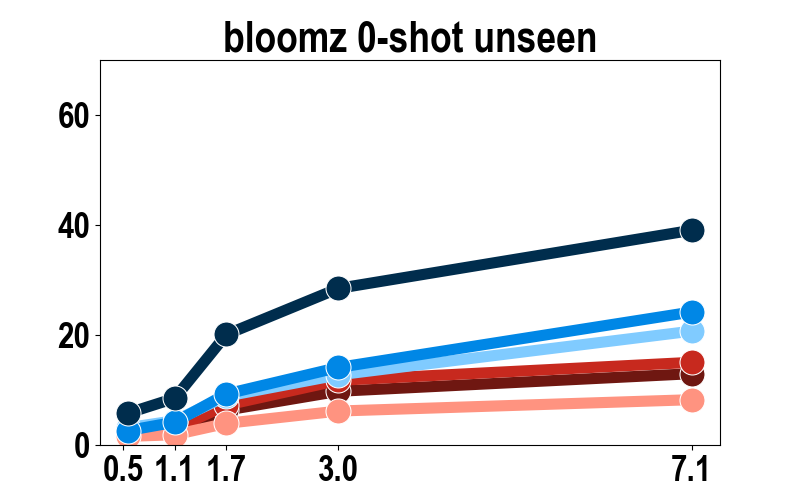}
            % \caption{}
            \label{fig:bloomz 0-shot_unseen_gen} 
        \end{subfigure}
    \end{minipage}%
    \begin{minipage}[t]{0.255\textwidth}
        \centering
        \begin{subfigure}[t]{0.9\textwidth}
            \includegraphics[width=\textwidth]{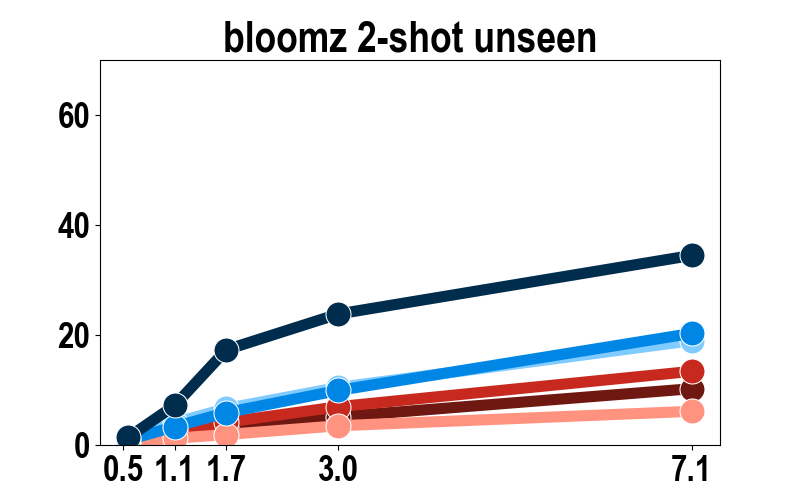}
            % \caption{}
            \label{fig:bloomz 2-shot_unseen_gen} 
        \end{subfigure}
    \end{minipage}
    
    \caption{{\bf Generation task}: Results of evaluation (SacreBLEU) across different models based on language resource level using Flores-200 dataset. The model sizes on the x-axis are in billions of parameters.
            \textcolor[HTML]{6f1711}{\rule{6pt}{6pt}} Resource Level 0, 
            \textcolor[HTML]{c7291e}{\rule{6pt}{6pt}} Resource Level 1, 
            \textcolor[HTML]{ff9380}{\rule{6pt}{6pt}} Resource Level 2, 
            \textcolor[HTML]{80cbff}{\rule{6pt}{6pt}} Resource Level 3, 
            \textcolor[HTML]{0087e6}{\rule{6pt}{6pt}} Resource Level 4, 
            \textcolor[HTML]{002d4d}{\rule{6pt}{6pt}} Resource Level 5.
    % \vspace{-0.3cm}
    }
    \label{fig:lineplots_class_gen}
\end{figure*}

\begin{table}[t]
    % \small
    \centering
    
    \begin{tabular}{c|c|c|c|c} 
    \toprule
    \multicolumn{5}{c}{\textbf{Text Classification} (F1 Score [0,1])}\\ 
    \midrule
    &0s seen & 0s unseen & 2s seen & 2s unseen\\ 
    \midrule
    \texttt{xglm} &0.0145 & 0.0071 & 0.0294 & {\bf -0.0050}\\ 
    \texttt{bloom} & 0.0002 & {\bf -0.0038} & 0.0561 & 0.0419\\ 
    \texttt{bloomz} & 0.0254 & 0.0126 & 0.0718 & 0.0502\\    
    \midrule \midrule
    \multicolumn{5}{c}{\textbf{Text Generation} ({\em Normalized} SacreBLEU Score [0,1])}\\ 
    \midrule
    &0s seen & 0s unseen & 2s seen & 2s unseen\\ \midrule
    \texttt{xglm} & 0.0254 & 0.0041 & 0.0054 & 0.0028\\ 
    \texttt{bloom} & 0.0101 & 0.0066 & 0.0154 & 0.0098\\
    \bottomrule
    \end{tabular}
        \caption{Slopes of linear fit for scaling data}
    \label{slopes}
    % \vspace{-0.2cm}
\end{table}

{We also find that the unseen languages are benefiting from related seen languages rather than their own resource levels. For example, in a classification task using both xglm-7.5b and bloomz-7b1 models, the unseen language Asturianu (F1 = 0.52), which is of resource level 1, performs comparably to its closely related seen language, Spanish (F1 = 0.56), which is at resource level 5. On the other hand, Croatian (F1 = 0.32), of resource level 4, performs worse than Asturianu, likely because it lacks closely related languages in the seen category. More analysis on resource level in Appendix.

\subsection{Comparing text classification vs. text generation} \quad To get a better  sense of the quality of the scaling for the classification and generation tasks, we provide the {\em slope} of the best linear fit for all the cases from Figure \ref{fig:zero_vs_few} in Table \ref{slopes}. By and large, we observe very small slopes in most cases for the generation task and significant positive slopes for the 2-shot case for the classification task. We believe that in the classification case, seeing examples helps the classification task by conditioning the probabilities to select one of the provided class choices. In the generation task, on the other hand, providing context appears to reduce the ability of the model to generalize and thus we see poorer scores for 2-shot. 
%(with the exception of \texttt{bloom}). 
{One possible reason for this issue could be the need for more few-shot examples in the generative task. However, exploring additional settings was beyond the scope of this research.} %\rp{uncommmented the last sentence}
 
Figure~\ref{fig:lineplots_class} presents the results for the \textbf{classification task} of \texttt{xglm} and \texttt{bloomz}, in 0-shot and 2-shot settings, further separated across seen and unseen languages, and the six different resource levels ({\texttt{bloom} plots in Appendix. Figure \ref{fig:lineplots_class_gen} presents the corresponding results for the \textbf{generative task}. Seen languages have a clear advantage over unseen languages in general. We observe that \texttt{xglm} treats languages from all resource levels fairly equally, whereas \texttt{bloom} and \texttt{bloomz} models show a considerable gap between the three higher levels (3, 4, 5) and the three lower levels (0, 1, 2). In the unseen scenario, the distinction appears not along high or low but between resource level 5 languages and the rest of the languages. The impact of 0- and 2-shot on seen/unseen languages is particularly noteworthy. In Section~\ref{sec:settings} we saw that few-shot ICL hurts smaller models. Here, we see that few-shot ICL disproportionately hurts smaller models' performance on lower-resource languages compared to higher-resource languages.

%% file: AnonymousSubmission/files/Sections/6_conclusion.tex
\section{Conclusion}

We explored scaling trends in multilingual contexts, using three multilingual LLMs and {two task types} covering 204 languages. Our results highlight the complexities of multilingual model performance and scalability, emphasizing the need for careful consideration of various factors including whether the languages were seen or not during pretraining and shot settings (whether zero-shot or few-shot). Recently, \citet{mosbach2023fewshot} showed that fine-tuning outperforms in-context learning in both in-domain and out-of-domain performance. On the other hand, another study shows that fine-tuning limits the generalization ability of the models introducing false correlations \citep{shliazhko2024mgpt}. In light of such studies, it would be interesting to explore the (positive or negative) effects of fine-tuning in extensive multilingual scenarios in future work.
\citet{cao2024GMM} highlighted that generative tasks differ from classification tasks. Investigating approaches to enhance the generalization of instruction-tuned models to unseen languages, considering that few-shot learning might not always be the optimal solution, presents a compelling direction for future research in natural language generation. %\rp{needs proof reading}

%\section*{Limitations}
%\rp{Shall we remove this section since the template does not need it?}
%Due to resource constraints, we were unable to evaluate the largest size of the two models (i.e. \texttt{bloom-176b} and \texttt{bloomz-176b}) in our experiments. Informed by prior work, we used English prompts (in 0- and 2-shot settings) but language-specific exemplars (in 2-shot settings). We did not tune the prompts specifically to any model, nor were we able to consider the effects of prompt sensitivity in multilingual scenarios. 

\section*{Ethics Statement}
{While we have included over 200 dialects and languages in our study, we acknowledge that many more languages remain to be comprehensively studied.}

%, which their developers claim to be the most efficient ones.} 

%% file: AnonymousSubmission/files/Sections/7_appendix.tex
% \subsection{Dataset}
% \label{app:dataset}

% Table \ref{tab:samples} shows examples from SIB-200 and Flores-200 datasets.

\section{0-shot vs. 2-shot settings}\label{sec:settings} 

\begin{figure}[t]
    \centering
    
    \begin{subfigure}[b]{0.226\textwidth}
        \includegraphics[width=\textwidth]{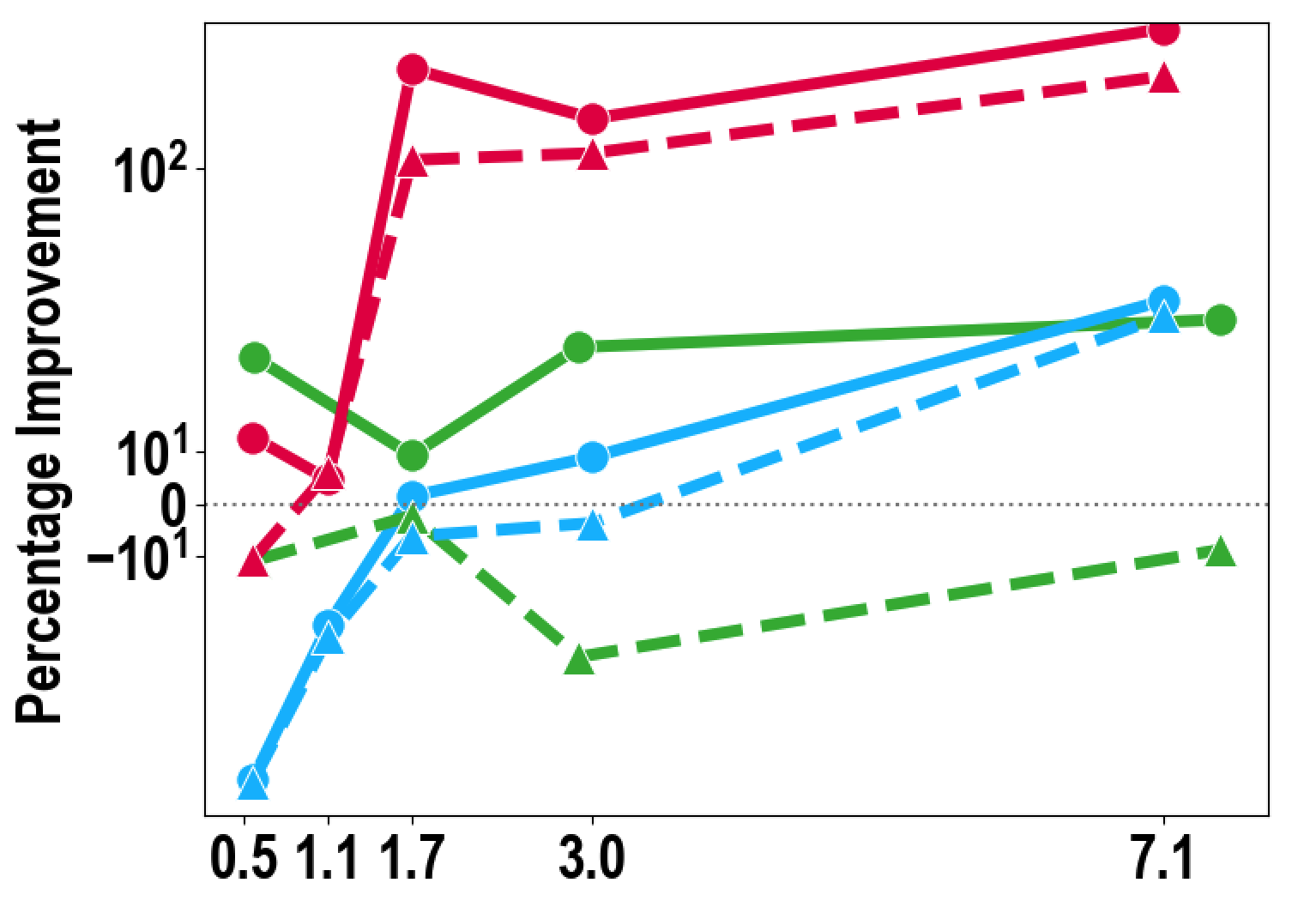}
        % \caption{Figure 1}
        % \label{fig:figure1}
    \end{subfigure}
    % \hfill
    \begin{subfigure}[b]{0.22\textwidth}
        \includegraphics[width=\textwidth]{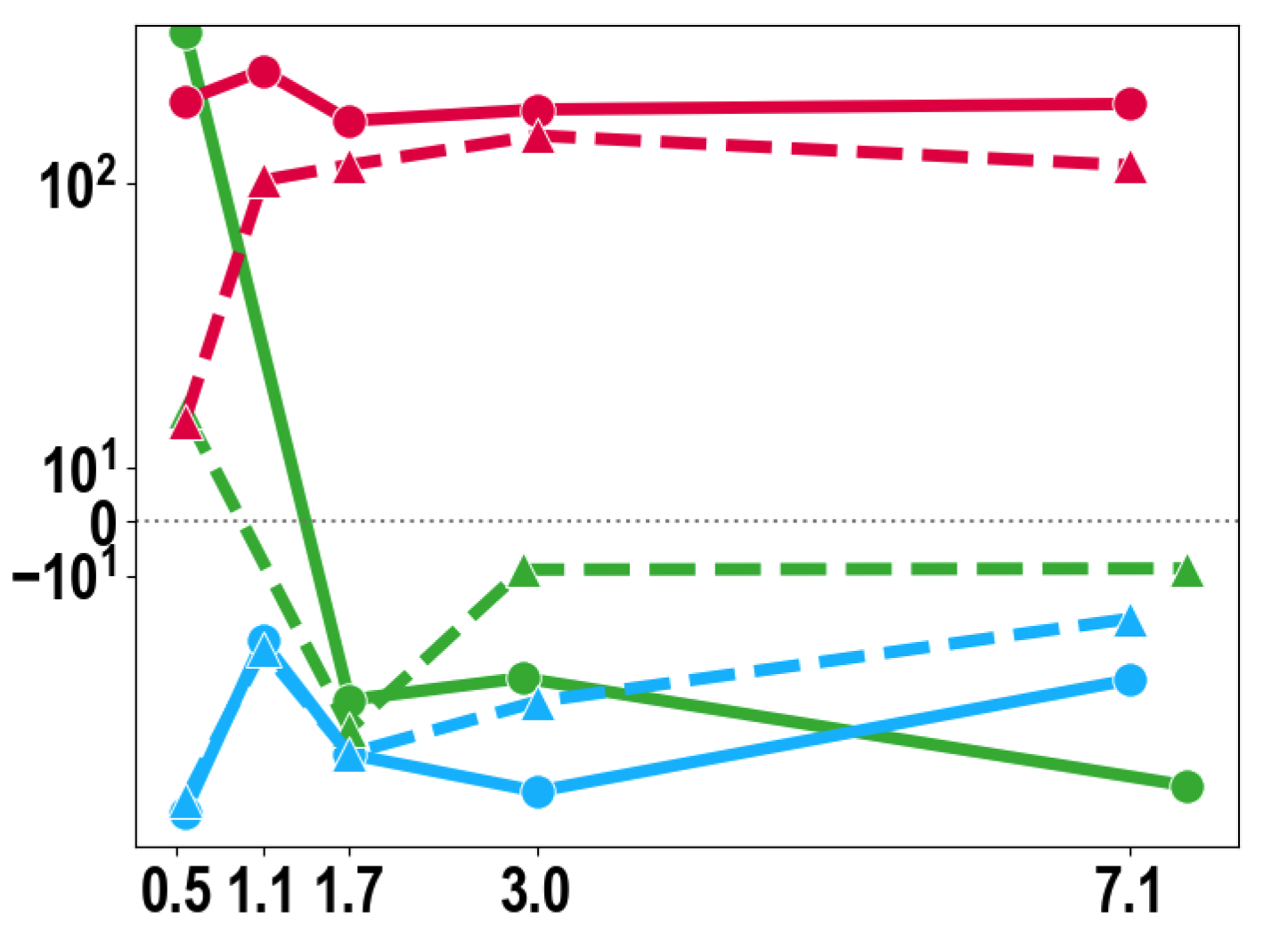}
        % \caption{Figure 2}
        % \label{fig:figure2}
    \end{subfigure}
    
    % \vskip 0.2\baselineskip
    
    \begin{subfigure}[b]{0.38\textwidth}
        \includegraphics[width=\textwidth]{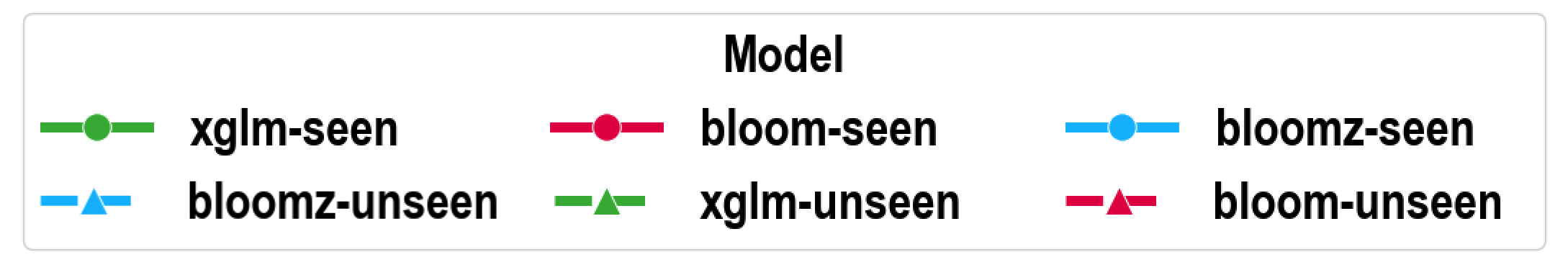}
        % \caption{Figure 3}
        % \label{fig:figure3}
    \end{subfigure}
    \caption{Relative improvement  in log-scale on y-axis from 0-shot to 2-shot. Text classification results are on the left, and on the right are text generation results.} \label{fig:delta_plot}
     % \vspace{-0.3cm}
\end{figure}

Figure ~\ref{fig:delta_plot} shows the relative improvement in percentage from 0-shot to 2-shot. For the {\bf classification task},  on the one hand, we find that in 5 out of 6 cases, the larger models make more efficient use of in-context information \cite{brownGPT3,weietal2023inverse}. On the other hand, in 4 out of 6 cases, we note that few-shot ICL {\em hurts} smaller models where the percentage increment is negative (less than 0). If we look at Table \ref{slopes} which provides the slope of the linear fit for the models, we observe that the slope for the 2-shot setting is higher for both seen and unseen languages and for all three models with the exception of \texttt{xglm}, for which performance degrades marginally with model size when going from 0-shot to 2-shot for unseen languages.

For the \textbf{generative task}, in Figure \ref{fig:delta_plot} (right) we observe that the performance of \texttt{bloom} improves in the 2-shot setting for seen as well as unseen languages whereas \texttt{xglm} and \texttt{bloomz} display the opposite behavior, suggesting that 2-shot setting does not seem to help in the generation task on this dataset.

\section{BLOOM Model Performance At Different Resource Levels} \label{app:bloom_resource_levels}
\begin{figure*}[t]
    \centering

    \begin{minipage}[t]{0.255\textwidth}
        \centering
        \begin{subfigure}[t]{0.9\textwidth}
            \includegraphics[width=\textwidth]{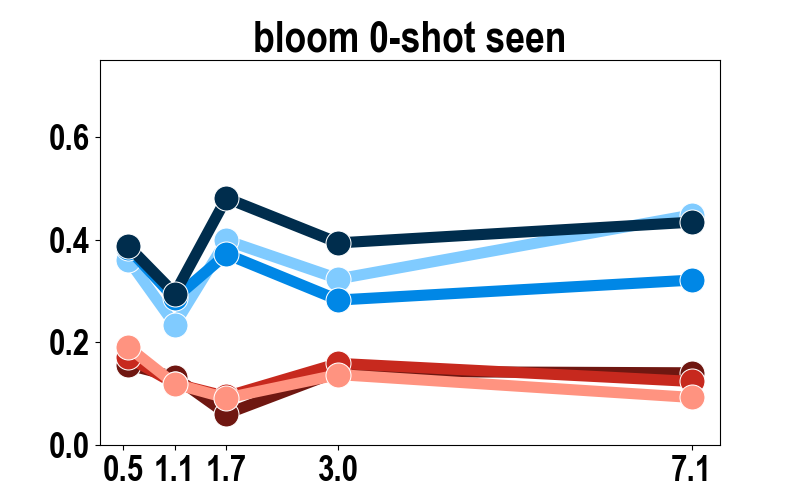}
            % \caption{}
            \label{fig:bloom 0-shot_seen_class} 
        \end{subfigure}
    \end{minipage}%
    \begin{minipage}[t]{0.255\textwidth}
        \centering
        \begin{subfigure}[t]{0.9\textwidth}
            \includegraphics[width=\textwidth]{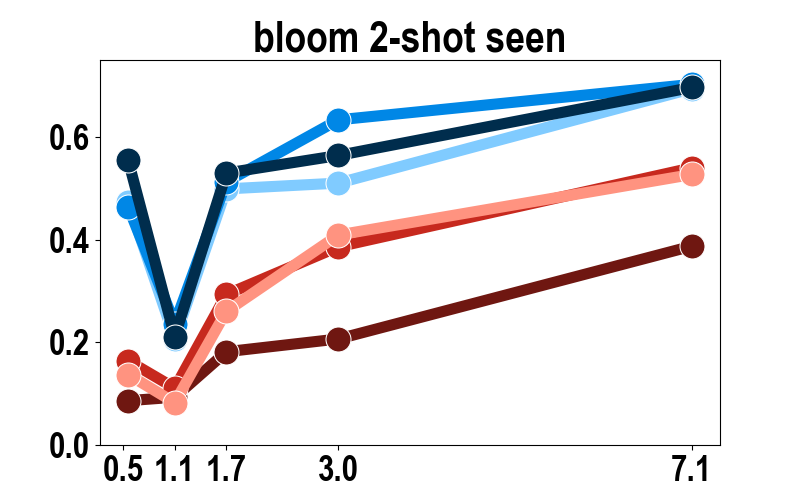}
            % \caption{}
            \label{fig:bloom 2-shot_seen_class} 
        \end{subfigure}
    \end{minipage}%
    \begin{minipage}[t]{0.255\textwidth}
        \centering
        \begin{subfigure}[t]{0.9\textwidth}
            \includegraphics[width=\textwidth]{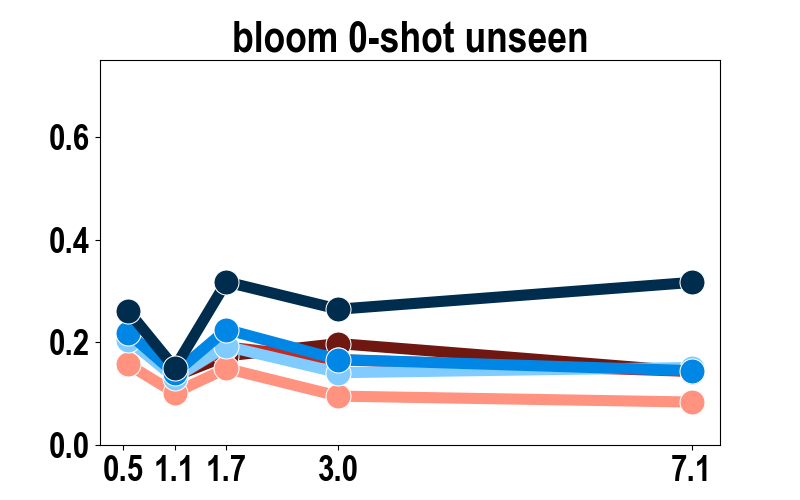}
            % \caption{}
            \label{fig:bloom 0-shot_unseen_class} 
        \end{subfigure}
    \end{minipage}%
    \begin{minipage}[t]{0.255\textwidth}
        \centering
        \begin{subfigure}[t]{0.9\textwidth}
            \includegraphics[width=\textwidth]{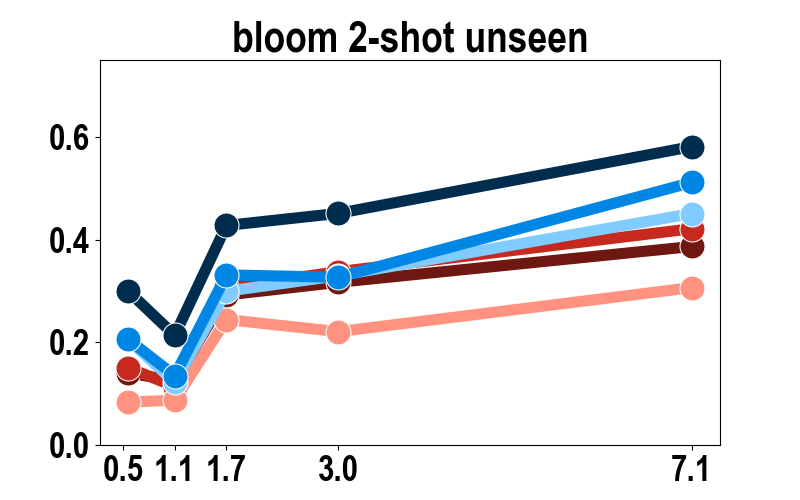}
            % \caption{}
            \label{fig:bloom 2-shot_unseen_class} 
        \end{subfigure}
    \end{minipage}
    
    \caption{Results of evaluation (F1 score) across different models based on language resource level for the \textbf{classification} task using SIB-200 dataset. The model sizes on the x-axis are in billions of parameters.
            \textcolor[HTML]{6f1711}{\rule{6pt}{6pt}} Resource Level 0, 
            \textcolor[HTML]{c7291e}{\rule{6pt}{6pt}} Resource Level 1, 
            \textcolor[HTML]{ff9380}{\rule{6pt}{6pt}} Resource Level 2, 
            \textcolor[HTML]{80cbff}{\rule{6pt}{6pt}} Resource Level 3, 
            \textcolor[HTML]{0087e6}{\rule{6pt}{6pt}} Resource Level 4, 
            \textcolor[HTML]{002d4d}{\rule{6pt}{6pt}} Resource Level 5.
    }
    \label{fig:lineplots_class_bloom}
    % \vspace{-0.5cm}
\end{figure*}

\begin{figure*}[t]
    \centering
    \begin{minipage}[t]{0.255\textwidth}
        \centering
        \begin{subfigure}[t]{0.9\textwidth}
            \includegraphics[width=\textwidth]{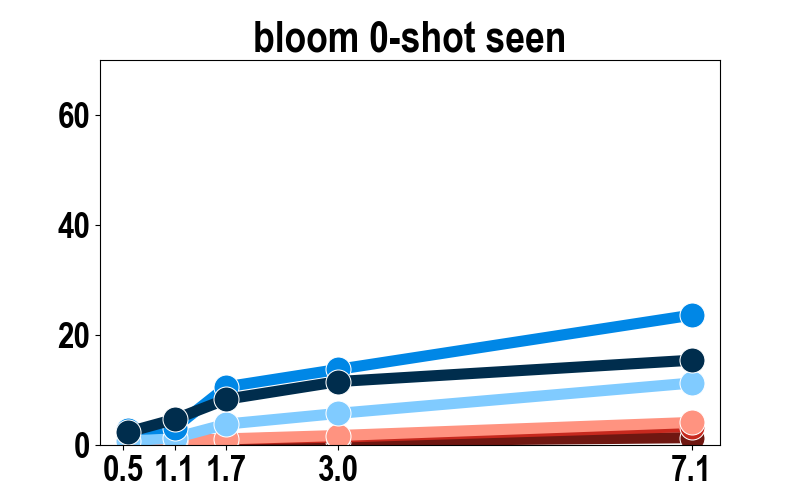}
            % \caption{}
            \label{fig:bloom 0-shot_seen_gen} 
        \end{subfigure}
    \end{minipage}%
    \begin{minipage}[t]{0.255\textwidth}
        \centering
        \begin{subfigure}[t]{0.9\textwidth}
            \includegraphics[width=\textwidth]{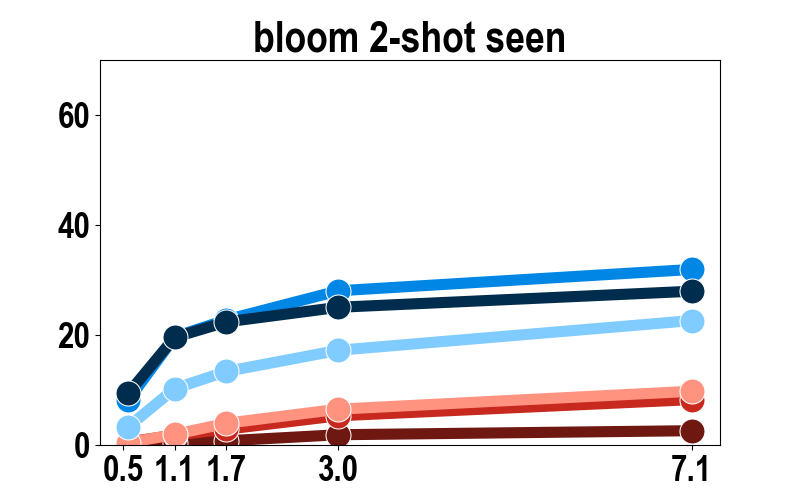}
            % \caption{}
            \label{fig:bloom 2-shot_seen_gen} 
        \end{subfigure}
    \end{minipage}%
    \begin{minipage}[t]{0.255\textwidth}
        \centering
        \begin{subfigure}[t]{0.9\textwidth}
            \includegraphics[width=\textwidth]{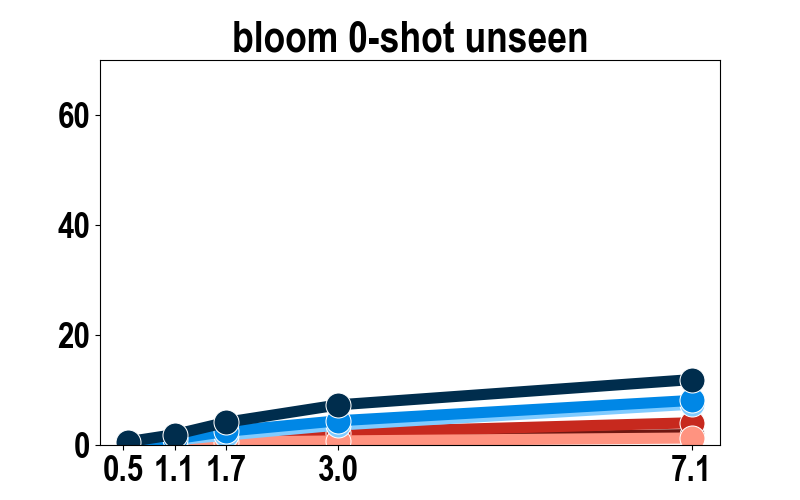}
            % \caption{}
            \label{fig:bloom 0-shot_unseen_gen} 
        \end{subfigure}
    \end{minipage}%
    \begin{minipage}[t]{0.255\textwidth}
        \centering
        \begin{subfigure}[t]{0.9\textwidth}
            \includegraphics[width=\textwidth]{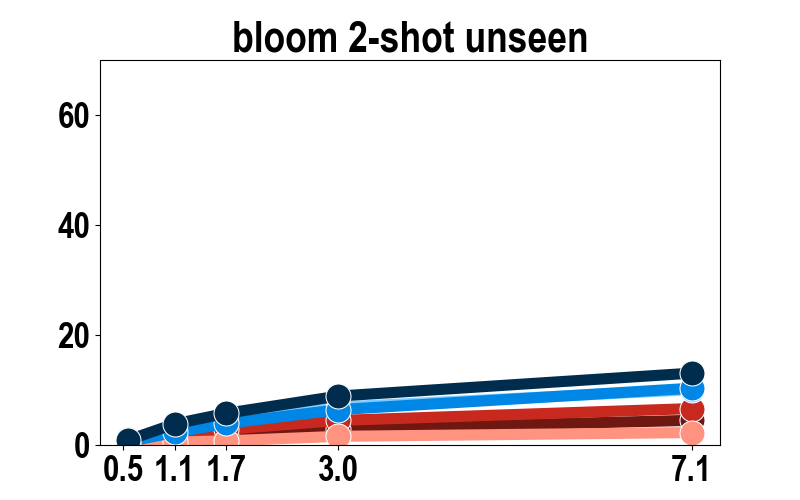}
            % \caption{}
            \label{fig:bloom 2-shot_unseen_gen} 
        \end{subfigure}
    \end{minipage}
    
    \caption{Results of evaluation (SacreBLEU) across different models based on language resource level for the \textbf{generation} task using Flores-200 dataset. The model sizes on the x-axis are in billions of parameters.
            \textcolor[HTML]{6f1711}{\rule{6pt}{6pt}} Resource Level 0, 
            \textcolor[HTML]{c7291e}{\rule{6pt}{6pt}} Resource Level 1, 
            \textcolor[HTML]{ff9380}{\rule{6pt}{6pt}} Resource Level 2, 
            \textcolor[HTML]{80cbff}{\rule{6pt}{6pt}} Resource Level 3, 
            \textcolor[HTML]{0087e6}{\rule{6pt}{6pt}} Resource Level 4, 
            \textcolor[HTML]{002d4d}{\rule{6pt}{6pt}} Resource Level 5.
    % \vspace{0.15cm}
    }
    \label{fig:lineplots_gen_bloom}
\end{figure*}

Figure~\ref{fig:lineplots_class_bloom} and ~\ref{fig:lineplots_gen_bloom} show the performance of BLOOM model on classification and generation task across different resource levels.

\section{Correlation Between Performance And Pretraining Data And Resource Levels} \label{app:correlation_pretrain_and_resource_level}

In order to better understand the scaling performance of the models, we dig in deeper into the languages that form the pretraining corpus. We experiment with using a rough proxy attribute such as general resource levels as introduced by \citet{joshi-etal-2020-state} where languages of the world are categorized into 6 subsets, with level `0' representing the cluster of languages with the lowest amounts of data available (i.e., extremely low resource languages), and level `5' representing the languages that enjoy considerably large amounts of unannotated and annotated data (i.e., high resource languages). We now analyze the correlation between model performance and properties of the languages in more detail by computing the correlation between model F1 score and pretraining data (PD), resource level (RL), and merged resource level (RL*) as illustrated in Table \ref{tab: zero_shot_corr} for the 0-shot case.

A prevalent hypothesis suggests that the performance of a model on a given language may be correlated with the amount of language-specific data present in the pretraining corpus \cite{imanigoogharietal2023glot500,adelani2024sib200}. %For instance, Pearson's correlation of Glot500-m's zero-shot performance on a sentence retrieval dataset is $r$ = 0.34. 
For the seen languages, we compute the Pearson's correlation between the models' F1 score and the pretraining data distribution as shown in the \textbf{second column} of Table~\ref{tab: zero_shot_corr}. \texttt{xglm} obtains a correlation of at most 0.18, \texttt{bloom} shows moderate correlation ($r$ = up to 0.68) whereas \texttt{bloomz} shows weak to moderate correlation ($r$ = up to 0.4). These results indicate that: 1) a model's performance is not always correlated with the amount of language-specific data as in the case of \texttt{xglm}, and 2) despite being trained on the same pretraining corpus, different sized models show different correlations.

We next consider the correlation between model performance and language resource levels (\textbf{third column} of Table~\ref{tab: zero_shot_corr}). In the context of the \textbf{classification task}, we can make several observations: 1) The strongest correlation between the performance of seen languages and their resource levels   is shown by \texttt{bloom}, followed \texttt{bloomz}, and lastly, \texttt{xglm}. This is reflected in how these models perform in the languages of the different resource levels. Our discussion carries forth to the \textbf{generation task} as well, with one notable exception. We see that \texttt{xglm} for the 2-shot case shows the highest performance for seen `0' resource languages. This is a statistical aberration because there is just one resource `0' language in the seen category for \texttt{xglm}. 

\begin{table*}[!t]
\centering
% \small
\begin{tabular}{p{2cm} p{0.85cm} p{0.85cm} p{0.85cm}| p{0.85cm}}
\toprule
\multicolumn{1}{c}{} & \multicolumn{2}{c}{\hspace{0.8cm} seen} & & \multicolumn{1}{c}{unseen} \\
\cmidrule{2-5}
Model &PD &RL &RL* &RL*\\
\midrule
\texttt{xglm-564m} &0.12 &0.17 &0.28 &0.12\\
\texttt{xglm-1.7b} &0.07 &0.02 &0.14 &0.17\\
\texttt{xglm-2.9b} &0.18 &0.14 &0.23 &0.16\\
\texttt{xglm-7.5b} &0.08 &0.03 &0.06 &0.06\\
\midrule
\texttt{bloom-560m} &0.44 &0.67 &0.71 &0.01\\
\texttt{bloom-1b1} &0.41 &0.64 &0.68 &0.04\\
\texttt{bloom-1b7} &0.53 &0.80 &0.84 &0.09\\
\texttt{bloom-3b} &0.68 &0.70 &0.74 &-0.04\\
\texttt{bloom-7b1} &0.43 &0.72 &0.83 &0.1\\
\midrule
\texttt{bloomz-560m} &0.40 &0.63 &0.66 &0.07\\
\texttt{bloomz-1b1} &0.30 &0.68 &0.68 &0.05\\
\texttt{bloomz-1b7} &0.28 &0.57 &0.59 &0.002\\
\texttt{bloomz-3b} &0.27 &0.61 &0.62 &0.06\\
\texttt{bloomz-7b1} &0.20 &0.59 &0.59 &0.16\\
\bottomrule
\end{tabular}
\caption{Pearson correlations using F1 score for classification task in 0-shot experiments and PD, RL, and RL*, where PD = Pretraining Data, RL = Resource Level, and RL* is after merging low resource levels (i.e. 0, 1, 2) and high resource levels (i.e. 3, 4, 5) into binary categories of low and high, respectively. Similar correlation were obtained for 2-shot. The PD for \texttt{xglm} models is the ratio with low resource upsampling from their original documentation.} 
\label{tab: zero_shot_corr}
% \vspace{-0.4cm}
\end{table*}

%Depending on language-specific training data shows us only a small snapshot of the overall picture, given that only about 14 to 22\% of the 204 languages of SIB-200 are `seen' by these models, leaving the remaining as `unseen'. How can we obtain correlation between model performance and unseen languages (which do not have pretraining data distributions)? 

%Therefore, we compute the correlation between F1 scores and general resource level (\textbf{third column}) and find that F1 scores are more strongly correlated with general resource levels than they were with language-specific pretraining data distribution possibly due to the noise introduced by a larger set of values in the latter. We also note that \texttt{bloom} and \texttt{bloomz} continue to exhibit strong correlation suggesting that their performance for higher resource and lower resource languages will be more prominently distinct (discussed later in the next section). 

Continuing this line of investigation, we merge the three lower resource levels into one, and the three higher resource levels into another, thus obtaining a binary categorization of low and high. Using this, we compute the correlation one more time between F1 and low/high resourcedness (\textbf{fourth column}) and find that the correlation becomes even stronger. 

Finally, we extend this analysis to unseen languages by computing the correlation of the F1 scores and the (merged) general resource availability (\textbf{fifth column}). Rather surprisingly, we notice that all models exhibit poor correlation. One possible explanation for this phenomenon is found in our analysis comparing the performance of seen and unseen languages from the same language family earlier in Figure~\ref{fig:f1_vs_family} where the results confirm the assessment that unseen languages are able to obtain performance closer to seen languages from the same language family, possibly leveraging some benefits of cross-lingual transfer, which is certainly not being captured by simple correlations involving pretraining data distributions or resource levels. 

Overall, contrary to the results in \cite{winata2022cross}, we show that a model’s performance is not always correlated with the amount of language-specific data. Instead, our results suggest that general resource level seems to be a stronger indicator of performance for seen languages (but not for unseen languages).

Future work could explore enhancing the performance of low-resource languages while expanding the representation of resource levels.